\documentclass[journal]{IEEEtran}

\usepackage{amsfonts}
\usepackage{amssymb}
\usepackage{amsmath}
\DeclareMathOperator{\diag}{diag}
\DeclareMathOperator*{\argmax}{arg\,max}

\usepackage{xspace}
\usepackage{intmacros}
\usepackage{subfigure}

\usepackage[final]{graphicx}

\newcommand{\Sb}{{\textbf{S}}}

\newcommand{\sbb}{{\textbf{s}}}
\newcommand{\wbb}{{\textbf{w}}}

\newcommand{\cb}{{\textbf{c}}}
\newcommand{\Cb}{{\textbf{C}}}
\newcommand{\Db}{{\textbf{D}}}

\newcommand{\bb}{{\textbf{b}}}

\newcommand{\Hb}{{\textbf{H}}}
\newcommand{\Ib}{{\textbf{I}}}

\newcommand{\Rb}{{\textbf{R}}}

\newcommand{\lone}{$\ell^{1}$}
\newcommand{\ltwo}{$\ell^{2}$}
\newcommand{\lp}{$\ell^{p}$}

\newcounter{examplecounter}
\renewcommand{\theexamplecounter}{\arabic{examplecounter}}

\ifCLASSINFOpdf
\else
\fi

\begin{document}
%
\title{Iterative Bayesian Reconstruction of Non-IID Block-Sparse Signals}
%
%
%

\author{Mehdi~Korki,~\IEEEmembership{Student Member,~IEEE,}
        Jingxin~Zhang,~\IEEEmembership{Member,~IEEE,}~Cishen~Zhang,~\IEEEmembership{Senior Member,~IEEE,}
        and~Hadi~Zayyani
\thanks{M. Korki, J. Zhang, and C. Zhang are with the Department
of Telecommunications, Electrical, Robotics and Biomedical Engineering, Swinburne University of Technology, Hawthorn,
3122 Australia (e-mail: mkorki@swin.edu.au; jingxinzhang@swin.edu.au; cishenzhang@swin.edu.au).}
\thanks{H. Zayyani is with the Department of Electrical and Computer Engineering, Qom University of Technology, Qom, Iran  (e-mail: zayyani2009@gmail.com).}} 
\maketitle

\begin{abstract}
This paper presents a novel Block Iterative Bayesian Algorithm (Block-IBA) for reconstructing block-sparse signals with unknown block structures. Unlike the existing algorithms for block sparse signal recovery which assume the cluster structure of the nonzero elements of the unknown signal to be independent and identically distributed 
(i.i.d.), we use a more realistic Bernoulli-Gaussian hidden Markov model (BGHMM) to characterize the non-i.i.d. block-sparse signals commonly encountered in practice. The Block-IBA iteratively estimates the amplitudes and positions of the block-sparse signal using the steepest-ascent based Expectation-Maximization (EM), and optimally selects the nonzero elements of the
block-sparse signal by adaptive thresholding. The global convergence of Block-IBA is analyzed and proved, and the effectiveness of Block-IBA is demonstrated by numerical experiments and simulations on synthetic and real-life data.
\end{abstract}


%
\IEEEpeerreviewmaketitle

\section{Introduction}
%
%
%
%
\IEEEPARstart{C}onsider the general Bayesian linear model
\begin{equation}
\label{equ_1}
\mathbf{y}=\mathbf{\Phi}\mathbf{w}+\mathbf{n}
\end{equation}
where $\mathbf{\Phi} \in \mathbb{R}^{N\times M}$ is a known measurement matrix,
$\mathbf{y} \in \mathbb{R}^{N}$ is the available measurement vector, and $\mathbf{n} \in \mathbb{R}^{N}$ is
the Gaussian corrupting noise. We aim to  estimate the original unknown signal $\mathbf{w} \in \mathbb{R}^{M}$ when $N\ll M$. Under this condition the underdetermined system of linear equations in (\ref{equ_1}) has an infinite number of solutions, which makes the problem challenging and requires appropriate prior knowledge about the
unknown signal $\mathbf{w}$.

The problems based on the general linear model in (\ref{equ_1}) frequently occur in the fields of signal processing, statistics, neuroscience and machine learning. Examples of common applications, among many others, include compressed sensing \cite{refe_27}, \cite{refe_28},
sparse component analysis (SCA) \cite{refe_29}, sparse representation \cite{refe_30}-\nocite{refe_31}\nocite{refe_32}\cite{refe_33}, source localization \cite{refe_34}, 
\cite{refe_35}, and in particular direction of arrival (DOA) estimation \cite{refe_36}. An appropriate prior knowledge that can lead to recovery of $\mathbf{w}$
is the sparsity, namely, the majority of the elements of the unknown vector $\mathbf{w}$ are zero (or near zero), while only a few 
components are nonzero. Knowing the sparsity of vector $\mathbf{w}$ {\it a priori}, a theoretically proven and practically effective approach to recover the signal $\mathbf{w}$ is to
solve the following optimization problem

\begin{equation}
\label{equ_opt}
\widehat{\mathbf{w}}=\argmin_\mathbf{w}\beta \left \| \mathbf{y}-\mathbf{\Phi \mathbf{w}} \right \|_{2}^{2}+\tau \left \| \mathbf{w} \right \|_{1}
\end{equation}
where $\tau$ is the regularization parameter that controls the degree of the sparsity of the solution. Moreover, the effect of Gaussian noise $\mathbf{n}$ with zero mean and variance 
$\beta^{-1}/2$ is implicitly embedded in (\ref{equ_opt}). Some popular optimization algorithms have been developed to solve (\ref{equ_opt}) \cite{refe_33}, \cite{refe_37}-
\nocite{refe_38}\cite{refe_39}. Moreover, in some works such as \cite{refe_38}, \cite{refe_39} the developed sparse reconstruction algorithms use \lp-norm to replace 
\lone-norm where $0< p\leq 1$.

Compressed sensing (CS) aims to recover the sparse signal 
from underdetermined systems of linear equations. If the structure of the
signal is exploited, the better recovery performance can be achieved. A block-sparse signal, 
in which the nonzero samples manifest themselves as clusters, is an important structured
sparsity. Block-sparsity has a wide range of applications in multiband signals~\cite{refe_1}, 
audio signals~\cite{refe_2}, structured
compressed sensing~\cite{refe_3}, and the multiple measurement vector (MMV) model~\cite{refe_4}. CS for block-sparse signals is to estimate the original unknown signal $\mathbf{w} \in \mathbb{R}^{M}$ with the cluster structure

\begin{equation}
\label{equ_2}
\mathbf{w}=[\underbrace{w_{1},\ldots,w_{d_{1}}}_{\mathbf{w}^{T}[1]},\ldots,\underbrace{w_{d_{g-1}+1},
\ldots,w_{d_{g}}}_{\mathbf{w}^{T}[g]}]^{T}
\end{equation}
where $\mathbf{w}[i]$ denotes the $i$th block with length $d_{i}$ which are not necessarily identical. In the
block partition (\ref{equ_2}), only $k\ll g$ vectors $\mathbf{w}[i]$ have nonzero Euclidean norm. 

Given the {\it a priori}
knowledge of
block partition, a few algorithms such as  Block-OMP \cite{refe_5}, mixed \ltwo/\lone norm-minimization \cite{refe_6}, group LASSO \cite{refe_7}
and model-based CoSaMP \cite{refe_8}, work effectively in the block-sparse signal recovery. 
These algorithms require the knowledge of
the block structure (e.g. the location and the lengths of the blocks) in (\ref{equ_2}). However, in many applications, such prior knowledge is often unavailable. For instance, the accurate tree structure of the coefficients for the clustered sparse representation of the images
is unknown {\it a priori}. 
The impulsive noise estimation in Power Line Communication (PLC) is often cast into a block-sparse signal reconstruction problem, where the impulsive 
noise (i.e. signal $\mathbf{w}$) occurs in bursts with unknown locations and lengths \cite{refe_64}, \cite{refe_40}.

To recover the structure-agnostic block-sparse signal, some algorithms, e.g. CluSS-MCMC \cite{refe_9}, BM-MAP-OMP \cite{refe_10}, Block Sparse Bayesian Learning (BSBL) \cite{refe_11}, and pattern-coupled SBL (PC-SBL) \cite{refe_12} have been proposed recently, which require less {\it a priori} information.
However, all these algorithms use the i.i.d. model to describe the cluster structure
of the nonzero elements of the unknown signal, which restricts their applicability and performance, see Section \ref{seca} for demonstrative examples. Because many practically important signals, e.g. the impulsive noise in PLC, do not satisfy the i.i.d. condition, it is necessary to develop reconstruction algorithms for block-sparse signals using a more realistic signal model. Also, in the above mentioned algorithms, there is a risk 
to choose unreliable support set of the signal $\mathbf{w}$, which may result in inappropriate sampling of nonzero elements of the signal $\mathbf{w}$. Hence, it is necessary to design an adaptive method to select the most probable support set based on the underlying structure of the signal. The ability of the algorithm to automatically tune up the signal
(i.e. $\mathbf{w}$) model parameters
is important, particularly when working with real-world datasets, but it is not provided by most of the existing block-sparse signal recovery algorithms (e.g., \cite{refe_9}-\nocite{refe_10}\nocite{refe_11}\cite{refe_12}).

To tackle the above mentioned problems, we propose a novel iterative Bayesian algorithm (Block-IBA) which

\begin{itemize}

\item uses a Bernoulli-Gaussian hidden Markov model (BGHMM)  \cite{refe_40} for the block-sparse signals. This 
model better captures the burstiness (block structure) of the impulsive noise and hence is more realistic for practical applications such as PLC

\item incorporates, different to the other algorithms \cite{refe_9}-\nocite{refe_10}\nocite{refe_11}\cite{refe_12}, an adaptive threshold technique for optimal selection of the columns of the sampling matrix $\mathbf{\Phi}$ to maximally sample the nonzero elements of signal $\mathbf{w}$. Using this technique, the Block-IBA improves the reconstruction performance for the block-sparse signals

\item uses a maximum {\it a posteriori} (MAP) estimation procedure to automatically
learn the parameters of the statistical signal model (e.g. the variance and the elements of state-transition matrix of BGHMM), averting complicated tuning updates.

\end{itemize}


The proposed Block-IBA reconstructs the supports and the
amplitudes of block-sparse signal $\mathbf{w}$ using an expectation maximization (EM) algorithm when its block structure is
completely unknown. In the expectation step (E-step) the
amplitudes of the signal $\mathbf{w}$ are estimated iteratively whereas
in the maximization step (M-step) the supports of the signal
$\mathbf{w}$ are estimated iteratively. To this end, we utilize a steepest-ascent algorithm after converting the estimation problem of
discrete supports to a continuous maximization problem. Although the steepest-ascent algorithm has been used in the literature for recovering the sparse signals (e.g. \cite{refe_21}), investigation of this method is unavailable in the literature of block-sparse signal recovery. 
As a result the proposed Block-IBA offers more reconstruction accuracy than the existing state-of-the-art algorithms for the non-i.i.d. block-sparse signals. This is verified on both synthetic and real-world signals, where the block-sparse signal comprises a large number of narrow blocks.

The rest of the paper is organized as follows. In Section \ref{sec_sigmodel}, we present the signal model. In Section \ref{sec_opt}, the optimum estimation of unknown signal $\mathbf{w}$ using MAP solution is 
proposed. Based on the MAP solution, a novel Block-IBA is developed in Section \ref{sec_Block-IBA}. The estimation of signal model parameters is presented in Section \ref{sec_params}. Section \ref{sec_converge} analyzes 
the global and local maxima properties of the Block-IBA. Experimental results are presented in Section \ref{sec_experiment}. Finally, conclusions are drawn in Section \ref{sec_conc}.

\subsection{Notation}
Lower-case letters (e.g., $x$) denote scalars. Boldfaced lower-case letters (e.g. $\mathbf{x}$), denote vectors, while boldfaced upper-case letters (e.g, $\mathbf{X}$) denote matrices. Sets are denoted by script notation (e.g., $\mathcal{S}$). The notations $\left ( \cdot \right )^{T}$ and $\hat{\left ( \cdot \right )}$ denote transpose and estimate, respectively. An $M$-by-$M$ identity matrix is denoted by $\boldsymbol{I}_{M}$. The probability density function (PDF) of a random variable $X$ is denoted by $p_{X}\left ( x \right )$, with subscript omitted when it is clear from the context. The Gaussian distribution with mean $\boldsymbol{b}$ and covariance matrix $\boldsymbol{C}$ is denoted by 
$\mathcal{N}\left(\boldsymbol{b},\boldsymbol{C}\right)$ and the PDF of a random variable $X$ corresponding to that distribution by $\mathcal{N}\left(x;\boldsymbol{b},\boldsymbol{C}\right)$. Finally, the expectation of a random variable is denoted by $\mathbb{E}\left \{ \cdot \right \}$.
\section{Signal Model}\label{sec_sigmodel}

In this paper, the linear model of (\ref{equ_1}) is considered as the measurement process. The measurement matrix $\mathbf{\Phi}$ is assumed known beforehand and also its
columns are normalized to have unit norms. Furthermore, we model the noise in model (\ref{equ_1}) as a stationary, additive white Gaussian noise (AWGN) process, with
$\mathbf{n}\sim \mathcal{N}\left(0,\sigma^{2}_{n}\boldsymbol{I}_{N}\right)$. To model the block-sparse sources ($\mathbf{w}$), we introduce two hidden random processes, $\mathbf{s}$ 
and
$\boldsymbol{\theta}$ \cite{refe_21}, \cite{refe_41}. The binary vector $\mathbf{s} \in \left\{0,1\right\}^{M}$ describes the support of $\mathbf{w}$, denoted $\mathcal{S}$, while the vector $\boldsymbol{\theta} \in 
\mathbb{R}^{M}$ represents the amplitudes of the active elements of $\mathbf{w}$. Hence, each element of the source vector $\mathbf{w}$ can be characterized as follows:
\begin{equation}
\label{equ_3}
w_{i}=s_{i}\cdot \theta_{i}
\end{equation}
where $s_{i}=0$ results in  $w_{i}=0$ and $i \notin \mathcal{S}$, while $s_{i}=1$ results in  $w_{i}=\theta_{i}$ and $i \in \mathcal{S}$. Hence, in vector form we can show that

\begin{equation}
\label{equ_vec} \mathbf{w}=\mathbf{S}\boldsymbol{\theta},\qquad
\mathbf{S}=\diag(\mathbf{s}) \in\mathbb{R}^{M \times M}
\end{equation}

To model the block-sparsity of the source vector $\mathbf{w}$, we assume that $\mathbf{s}$ is a stationary first-order Markov process defined by two transition probabilities:
$p_{10}\triangleq \mathrm{Pr}\left\{s_{i+1}=1|s_{i}=0\right\}$ and $p_{01}\triangleq \mathrm{Pr}\left\{s_{i+1}=0|s_{i}=1\right\}$ \cite{refe_41}. Moreover, it can be shown that in the steady 
state we have the following relation between the transition probabilities and the probabilities in a given state:

\begin{equation}
\label{equ_4}
\mathrm{Pr}\left\{s_{i}=0\right\}= p=\frac{p_{01}}{p_{10}+p_{01}}
\end{equation}

\begin{equation}
\label{equ_5}
\mathrm{Pr}\left\{s_{i}=1\right\}=1- p=\frac{p_{10}}{p_{10}+p_{01}}
\end{equation}
Therefore, the two parameters $p$ and $p_{10}$ completely describe the state process of the Markov chain. As a result, the remaining transition probability can be determined as
$p_{01}=\frac{p\cdot p_{10}}{(1-p)}$.
The length of the blocks of the block-sparse signal is determined by parameter $p_{01}$, namely, the average number of consecutive samples of ones is specified by $1/p_{01}$ in the
Markov chain. Note that the amplitude vector $\boldsymbol{\theta}$ has also a Gaussian distribution with $\boldsymbol{\theta}\sim \mathcal{N}\left(0,\sigma^{2}_{\theta}\boldsymbol{I}_{M}\right)$.
Therefore, from (\ref{equ_3}) it is obvious that $p\left(w_{i}\vert s_{i},\theta_{i}\right)=\delta(w_{i}-s_{i}\theta_{i})$, where $\delta(\cdot)$ is the Dirac delta function. Removing $s_{i}$ and $\theta_{i}$ by the 
marginalization rule, we can find the PDF of the sources as

\begin{equation}
\label{equ_7}
p(w_{i})=p\delta(w_{i})+(1-p)\mathcal{N}\left(w_{i};0,\sigma^{2}_{\theta}\right),
\end{equation}
where $\sigma^{2}_{\theta}$ is the variance of $\boldsymbol{\theta}$. Equation (\ref{equ_7}) shows that the distribution of the sources is a Bernoulli-Gaussian hidden Markov model (BGHMM) which is utilized to implicitly
express
the  block sparsity of the signal model due to the point-mass distribution at $w_{i}=0$ and the hidden variables $s_{i}$. In many communication systems such as PLC, the additive noise 
is highly impulsive, where the peak noise amplitudes reach up to 50 dB above the AWGN (or background noise) level \cite{refe_40}. In addition, the impulsive noise in PLC shows the 
bursty (clustered) nature \cite{refe_40}, \cite{refe_62} with samples no longer i.i.d.

Unlike the memoryless models such as Bernoulli-Gaussian model~\cite{refe_13}-\nocite{refe_15}\nocite{refe_16}\cite{refe_19} which consider the impulsive noise samples to be i.i.d.,
the BGHMM  \cite{refe_40}, \cite{refe_book1}, \cite{refe_63} with the first-order Markov chain model allows to better describe the typical bursty nature of impulsive noise with non-i.i.d. samples. It is well known that the power spectral density (PSD) of i.i.d. signals is wide band and flat. In contrast, for non-i.i.d. signals, the PSD is narrow band and spiky. Hence, the bandwidth and shape of PSD indicate a signal's closeness to or distance from being i.i.d. As shown in Section \ref{seca}, the parameter $p_{01}$ of BGHMM directly controls the bandwidth and shape of the PSD of a Block-sparse signal. The larger the $p_{01}$, the narrower the bandwidth and vice versa.

It is observed from (\ref{equ_7}) that the development of the 
amplitude vector 
$\boldsymbol{\theta}$ is
independent of the sparsity of the random process, $\mathbf{s}$. Hence, some of the amplitudes are pruned out by inactive coefficients (those which are associated with  $s_{i}=0$). In
fact, the nonzero amplitudes $\theta_{i}$ are the results of the amplitudes of $w_{i}$ conditioned on  $s_{i}=1$. Although higher-order Markov processes and/or more complex mixture of
Gaussian model can be utilized within the framework of Block-IBA, we focus on the first-order Markov processes and Bernoulli-Gaussian model to reduce
the complexity in the development of the algorithm.

\section{Optimum Estimation of $\mathbf{w}$}\label{sec_opt}


To obtain the optimum estimate of $\mathbf{w}$, we pursue a MAP approach. Hence, we first determine the MAP estimate of $\mathbf{s}$ which maximizes the posterior probability 
$p(\mathbf{s}|\mathbf{y})$. After estimating $\mathbf{s}$, the estimation of unknown original signal  $\mathbf{w}$ can be obtained by the estimation of $\boldsymbol{\theta}$.

\subsection{MAP Estimation of $\mathbf{s}$}

Using the Bayes' rule, we can rewrite $p(\mathbf{s}|\mathbf{y})$ as

\begin{equation}
\label{equ_8}
p(\mathbf{s}|\mathbf{y})=\frac{p(\mathbf{s})p(\mathbf{y}|\mathbf{s})}{\sum_{s}p(\mathbf{s})p(\mathbf{y}|\mathbf{s})}
\end{equation}
where the summation is over all the possible $\mathbf{s}$ vectors describing the support of $\mathbf{w}$. Note that the denominator in (\ref{equ_8}) is common to all posterior
likelihoods, $p(\mathbf{s}|\mathbf{y})$, and thus can be ignored as it is a normalizing constant. To evaluate $p(\mathbf{s})$, we know that the $\mathbf{s}$ vector is a stationary first-order
Markov process with two transition probabilities given in Section \ref{sec_sigmodel}. Therefore, $p(\mathbf{s})$ is given by

\begin{equation}
\label{equ_9}
p\left(\mathbf{s}\right)=p(s_{1})\prod^{M-1}_{i=1}p\left(s_{i+1}|s_{i}\right)
\end{equation}
where $p\left(s_{1}\right)=p^{(1-s_{1})}(1-p)^{s_{1}}$ and

\begin{equation}
\label{equ_10}
p\left(s_{i+1}|s_{i}\right)=
\begin{cases}
(1-p_{10})^{(1-s_{i+1})}p_{10}^{s_{i+1}} & \textrm{if} \quad s_{i} = 0 \\ \\
p_{01}^{(1-s_{i+1})}(1-p_{01})^{s_{i+1}} & \textrm{if} \quad  s_{i} = 1
\end{cases}
\end{equation}

It remains to calculate $p(\mathbf{y}|\mathbf{s})$. As $\mathbf{w}|\mathbf{s}$ is Gaussian, $\mathbf{y}$ is also Gaussian with zero mean and the covariance

\begin{equation}
\label{equ_11}
\boldsymbol{\Sigma_{\mathbf{s}}}=E\left[\mathbf{y}\mathbf{y}^{T}|\mathbf{s}\right]=\sigma^{2}_{n}\boldsymbol{I}_{N}+\sigma^{2}_{\theta}\boldsymbol{\Phi} 
\mathbf{S} \boldsymbol{\Phi}^{T}
\end{equation}
where $\mathbf{S}=\diag(\mathbf{s})$ as defined in (\ref{equ_vec}). Therefore, up to an inessential multiplicative constant factor ($\frac{1}{\pi^{N}}$), we can write the likelihood function as

\begin{equation}
\label{equ_12}
p\left(\mathbf{y}|\mathbf{s}\right)=\frac{\exp\left(-\frac{1}{2}\mathbf{y}^{T}{\boldsymbol{\Sigma}^{-1}_{\mathbf{s}}}\mathbf{y}\right)}{\det(\boldsymbol{\Sigma}_{\mathbf{s}})}
\end{equation}
Hence, the MAP estimate of $\mathbf{s}$ is given by

\begin{equation}
\label{equ_13}
\mathbf{s}_{\mathrm{MAP}}=\argmax_\mathbf{s}p(\mathbf{s})p(\mathbf{y}|\mathbf{s})
\end{equation}
where $p(\mathbf{s})$ is calculated using (\ref{equ_9}) and  (\ref{equ_10}), whereas the prior likelihood $p(\mathbf{y}|\mathbf{s})$ is given by (\ref{equ_12}). The maximization is
performed over all $2^M$ possible sets of $\mathbf{s}$ vectors.

\subsection{MAP Estimation of $\boldsymbol{\theta}$ using Gamma Prior}\label{M_E_theta}

After the binary vector $\mathbf{s}$ is estimated, we complete the estimation of the original unknown signal $\mathbf{w}$ by estimating the amplitude samples of the
$\boldsymbol{\theta}$ vector. To this end, we estimate the amplitudes with considering hyperprior over the inverse of the variance. Full details are given below.





Following the Sparse Bayesian Learning (SBL) framework \cite{refe_22}, we consider a Gaussian prior distribution for amplitude vector $\boldsymbol{\theta}$:

\begin{equation}
\label{equ_15}
p\left(\boldsymbol{\theta};\gamma_{i}\right) \sim \mathcal{N}\left(0,\boldsymbol{\Sigma}^{-1}_{0}\right)
\end{equation}
where $\boldsymbol{\Sigma} _{0}=\diag( \gamma_{1},  \gamma_{2},\cdots , \gamma_{M})$.
Furthermore, $\gamma_{i}$ are the non-negative elements of the hyperparameter vector $\boldsymbol{\gamma}$, that is $\boldsymbol{\gamma}\triangleq\left \{ \gamma _{i} \right \}$.
Based on the SBL framework, we use Gamma distributions as hyperpriors over the hyperparameters $\left \{ \gamma _{i} \right \}$:

\begin{equation}\nonumber
\label{equ_17}
p\left ( \boldsymbol{\gamma}  \right )= \prod_{i=1}^{M}\mathrm{Gamma}\left ( \gamma _{i}\mid a,b \right )= \prod_{i=1}^{M}\Gamma \left ( a \right )^{-1}b^{a}\gamma_{i} ^{a-1}e^{-b\gamma_{i} }
\end{equation}
where $\Gamma \left ( a \right )= \int_{0}^{\infty}t^{a-1}e^{-t}dt$ is the Gamma function. To obtain non-informative Gamma priors, we assign very small values, e.g. $10^{-4}$ to two
parameters $a$ and $b$. From (\ref{equ_vec}), we can rewrite the linear model of (\ref{equ_1}) as

\begin{equation}
\label{equ_18}
\mathbf{y}=\mathbf{\Phi}\mathbf{S}\boldsymbol{\theta}+\mathbf{n}=\boldsymbol{\Psi}\boldsymbol{\theta}+\mathbf{n}
\end{equation}
where $\boldsymbol{\Psi}=\mathbf{\Phi}\mathbf{S}$. Therefore, from the linear model of (\ref{equ_18}) and given the support vector $\mathbf{s}$, the likelihood function also has Gaussian 
distribution:

\begin{equation}
\label{equ_19}
p\left ( \mathbf{y}\mid \boldsymbol{\theta};\sigma ^{2}_{n}  \right )\sim \mathcal{N}_{y\mid \theta}\left ( \boldsymbol{\Psi}\boldsymbol{\theta} ,\sigma ^{2}_{n}\boldsymbol{I}_{N} \right )
\end{equation}
Using the Bayes' rule the posterior approximation of $\boldsymbol{\theta}$ is found as a multivariate Gaussian:

\begin{equation}
\label{equ_20}
p\left ( \boldsymbol{\theta}\mid \mathbf{y};\boldsymbol{\gamma },\sigma ^{2}_{n} \right )\sim \mathcal{N}_{\theta}\left ( \boldsymbol{\mu} _{\theta},\boldsymbol{\Sigma}_{\theta} \right )
\end{equation}
with parameters

\begin{equation}
\label{equ_21}
\boldsymbol{\mu} _{\theta}= \sigma ^{-2}_{n}\boldsymbol{\Sigma }_{\theta}\boldsymbol{\Psi}^{T} \mathbf{y}
\end{equation}

\begin{align}
\label{equ_22_1}
\boldsymbol{\Sigma }_{\theta} &= \left ( \sigma ^{-2}_{n}\boldsymbol{\Psi }^{T}\boldsymbol{\Psi } +\boldsymbol{\Sigma }_{0}\right )^{-1}\\
\label{equ_22_2}
& =\boldsymbol{\Sigma }_{0}^{-1}-\boldsymbol{\Sigma }_{0}^{-1}\boldsymbol{\Psi }^{T}\left ( \sigma ^{2}_{n}\boldsymbol{I}_{N}+\boldsymbol{\Psi }\boldsymbol{\Sigma }_{0}^{-1}\boldsymbol{\Psi }^{T} \right )^{-1}\boldsymbol{\Psi }\boldsymbol{\Sigma }_{0}^{-1}.
\end{align}

Therefore, given the hyperparameters $\gamma_{i}$ and noise variance $\sigma^{2}_{n}$, the MAP estimate of $\boldsymbol{\theta}$ is

\begin{align}
\label{equ_23_1}
\widehat{\boldsymbol{\theta }}_{MAP}=\boldsymbol{\mu }_{\theta }&=\left ( \mathbf{\Psi }^{T}\mathbf{\Psi } +\sigma^{2}_{n}\mathbf{\Sigma }_{0} \right ) ^{-1}\mathbf{\Psi }^{T}\mathbf{y}\\
\label{equ_23_2}
&=\mathbf{\Sigma }^{-1}_{0} \mathbf{\Psi }^{T}\left ( \mathbf{\Psi }\boldsymbol{\Sigma }^{-1}_{0}\mathbf{\Psi }^{T}+\sigma^{2} _{n}\boldsymbol{I}_{N} \right )^{-1}\mathbf{y}
\end{align}
where (\ref{equ_23_2}) follows the identity equation $\left ( \mathbf{A}+\mathbf{B}\mathbf{B}^{T} \right )^{-1}\mathbf{B}=\mathbf{A}^{-1}\mathbf{B}\left ( \mathbf{I}+\mathbf{B}^{T}\mathbf{A}^{-1} \mathbf{B}\right ) ^{-1}$, and $\boldsymbol{\Sigma} _{0}=\diag( \gamma_{1},  \gamma_{2},\cdots , \gamma_{M})$. Moreover, the
hyperparameters  $\gamma_{i}$ control the sparsity of the amplitudes $\theta_{i}$. Sparsity in the samples of the amplitudes occur when particular variables 
$\gamma_{i}\rightarrow \infty$, whose effect forces the $\mathit{i}$th sample to be pruned out from the amplitude estimate.
\footnote{In practice, we observe that when the estimates $\gamma_{i}$ become very large, e.g. $10^{5}$ so that the coefficient of the $\mathit{i}$th sample is numerically
indistinguishable from zero, then the associated sample in $\boldsymbol{\theta}$ is set to zero.}
To calculate $\widehat{\boldsymbol{\theta }}_{MAP}$, we can also use (\ref{equ_21}) directly in which we have two options for obtaining the covariance matrix
$\boldsymbol{\Sigma }_{\theta}$ using (\ref{equ_22_1}) and (\ref{equ_22_2}). Note that, the computational complexity for estimation of
$\boldsymbol{\Sigma }_{\theta}$ is different in (\ref{equ_22_1}) and (\ref{equ_22_2}). An $M\times M$ matrix inversion is required using (\ref{equ_22_1}), whereas an $N\times N$
matrix inversion is needed in  (\ref{equ_22_2}).

When the noise variance ($\sigma ^{2}_{n}$) is also unknown, we can place conjugate gamma prior on the inverse of the variance (i.e. $\beta \triangleq\sigma ^{-2}_{n}$) as $p\left ( \beta  \right )= \mathrm{Gamma}\left ( \beta \mid c,d \right )$, where $c=d=10^{-4}$.
In fact, the complexity of posterior distribution will be alleviated by using conjugate priors. To estimate the hyperparameters, we utilize the 
Relevance Vector Learning (RVL) which is maximization of the product of the marginal likelihood (Type-II maximum likelihood) and the priors over the hyperparameters 
$\boldsymbol{\gamma}$ and  $\beta$ ($\sigma ^{2}_{n}$)~\cite{refe_22}. Given the priors, the likelihood of the observations can be given as

\begin{equation}
\label{equ_25}
\begin{split}
p\left ( \mathbf{y}\mid \boldsymbol{\gamma },\beta,a,b,c,d \right )&=\mathcal{N}\left ( \mathbf{y}\mid 0,\mathbf{\Psi }\mathbf{\Sigma }^{-1}_{0}\mathbf{\Psi }
+\beta ^{-1}\boldsymbol{I}_{N} \right )\\
  &\quad\times p\left ( \mathbf{\gamma } \right )\times p\left ( \beta  \right )
\end{split}
\end{equation}
A maximum likelihood (ML) estimator which maximizes (\ref{equ_25}) can be used to find the unknown hyperparameter $\boldsymbol{\gamma}$ and $\beta$. To this end, we use
expectation maximization (EM) to compute the unknown variables iteratively. Hence, to compute the ML estimate of the unknown hyperparameters $\boldsymbol{\gamma}$ and $\beta$,
we treat $\boldsymbol{\theta}$ as the latent variables and apply the EM algorithm. Moreover, we define $\Theta \triangleq\left \{ \boldsymbol{\gamma },\beta  \right \}$ for brevity.
 The EM algorithm proceeds by maximizing the following expression

\begin{equation}
\label{equ_26}
\begin{split}
Q\left ( \boldsymbol{\Theta } \right )&=\mathbb{E}_{\boldsymbol{\theta }\mid \mathbf{y},\mathbf{\Theta }^{\left ( k \right )}}\left [ \mathrm{log}(p\left ( \mathbf{y}\mid\boldsymbol{\theta },\beta  \right ) p\left ( \boldsymbol{\theta }\mid\boldsymbol{\gamma }\right )p\left ( \boldsymbol{\gamma } \right )p\left ( \beta  \right ))\right ]\\
&= \mathbb{E}_{\boldsymbol{\theta }\mid \mathbf{y},\mathbf{\Theta }^{\left ( k \right )}}\left [ \mathrm{log}(p\left ( \boldsymbol{\gamma } \right )p\left ( \boldsymbol{\theta }\mid \boldsymbol{\gamma } \right ))\right ]\\
&\quad+\mathbb{E}_{\boldsymbol{\theta }\mid \mathbf{y},\mathbf{\Theta }^{\left ( k \right )}}\left [ \mathrm{log}(p\left ( \beta  \right )p\left ( \mathbf{y }\mid \boldsymbol{\theta },\beta  \right )) \right ]
\end{split}
\end{equation}
where $\mathbf{\Theta }^{\left ( k \right )}$ refers to the current estimate of hyperparameters. To estimate $\boldsymbol{\Theta}$, we observe that the first and second summand in
(\ref{equ_26}) are independent of each other. Hence, the estimate of $\boldsymbol{\gamma}$ and $\beta$ is separated into two different optimization problems. For obtaining
$\boldsymbol{\gamma}$, the following iterative expression can be solved

\begin{equation}
\label{equ_27}
\boldsymbol{\gamma }^{\left ( k+1 \right )}=\argmax_\mathbf{\boldsymbol{\gamma }}\mathbb{E}_{\boldsymbol{\theta }\mid \mathbf{y},\mathbf{\Theta }^{\left ( k \right )}}\left [ \mathrm{log}(p\left ( \boldsymbol{\gamma } \right )p\left ( \boldsymbol{\theta }\mid \boldsymbol{\gamma } \right )) \right ]
\end{equation}
Therefore, an update for hyperparameter $\boldsymbol{\gamma}$ by computing the first derivative of the first summand of (\ref{equ_26}) with respect to $\boldsymbol{\gamma}$ can be expressed as

\begin{equation}
\label{equ_28}
\gamma ^{\left ( k+1 \right )}_{i}=\frac{1+2a}{\left ( \mu _{\theta ,i} ^{\left ( k \right )}\right )^{2}+\Sigma ^{\left ( k \right )}_{\theta ,ii}+2b}
\end{equation}
where $\mu _{\theta ,i}$ denotes the $i$th entry of $\boldsymbol{\mu} _{\boldsymbol{\theta}}$ in (\ref{equ_21}) and $\Sigma_{\theta ,ii}$ denotes the $i$th diagonal element of the 
covariance matrix $\boldsymbol{\Sigma} _{\boldsymbol{\theta}}$ in (\ref{equ_22_1}) or  (\ref{equ_22_2}).

Following the same method, we need to solve the following optimization problem to estimate $\beta$

\begin{equation}
\label{equ_29}
\beta^{\left ( k+1 \right )}=\argmax_\mathbf{\beta}\mathbb{E}_{\boldsymbol{\theta }\mid \mathbf{y},\mathbf{\Theta }^{\left ( k \right )}}\left [ \mathrm{log}(p\left ( \beta \right )p\left ( \mathbf{y }\mid \boldsymbol{\theta },\beta \right )) \right ]
\end{equation}
Hence, the $\beta$ learning rule is calculated by setting the first derivative of the second summand in (\ref{equ_26}) with respect to $\beta$ to zero, resulting in

\begin{equation}
\label{equ_30}
\begin{split}
\frac{1}{\beta ^{\left ( k+1 \right )}}&=\frac{1}{N+2c}\left \{ \left \| \mathbf{y}-\boldsymbol{\Psi }\boldsymbol{\mu }_{\boldsymbol{\theta }}^{\left ( k \right )}\vphantom{\frac{1}{2}} \right \|_{2}^{2}\right.\\
&\quad+
\left.
\left (\beta ^{\left (k\right )} \right )^{-1}\sum_{i=1}^{M}\left [ 1-\gamma ^{\left ( k \right )}_{i}\Sigma ^{\left ( k \right )}_{\theta ,ii} \right ]+2d \right \}
\end{split}
\end{equation}

Having estimated the posterior probability of $\mathbf{s}$ and MAP estimate of amplitude vector $\boldsymbol{\theta}$, the estimation of unknown original signal $\mathbf{w}$ is
complete. However, the evaluation of (\ref{equ_13}) over all $2^M$ possible sets of $\mathbf{s}$ vectors is a computationally daunting task  when $M$ is large. The difficulty of this 
exhaustive search is obvious from (\ref{equ_8})-(\ref{equ_13}). Hence, in the following section, we propose an Iterative Bayesian Algorithm referred to as Block-IBA which reduces the
complexity of the exhaustive search.

\section{Block Iterative Bayesian Algorithm}\label{sec_Block-IBA}

Finding the solution for (\ref{equ_13}) through combinatorial search is computationally intensive. This is because the computation should be done over the discrete space. One way
around this exhaustive search is to convert the maximization problem into a continuous form. Therefore, in this section we propose a method to convert the problem into a continuous
maximization and apply a steepest-ascent algorithm to find the maximum value. To this end, we model the elements of $\mathbf{s}$ vector as a Gaussian Mixture (GM) with two
Gaussian variables centered around $0$ and $1$ with sufficiently small variances. Hence, each discrete element of  $\mathbf{s}$ vector, i.e. $s_{i}$ can be given as

\begin{equation}
\label{equ_31}
p(s_{1})\approx p~\mathcal{N}\left ( 0,\sigma ^{2}_{0} \right )+\left ( 1-p \right )\mathcal{N}\left ( 1,\sigma ^{2}_{0} \right )
\end{equation}
Moreover, the other elements of $\mathbf{s}$ vector, i.e. $s_{i+1}$ ($i=1,\cdots ,M-1$) can be expressed as

\begin{equation}
\label{equ_32}
p(s_{i+1})\approx
\begin{cases}
(1-p_{10})\mathcal{N}\left ( 0,\sigma ^{2}_{0} \right )+p_{10} \mathcal{N}\left ( 1,\sigma ^{2}_{0} \right ) & \textrm{if} \quad s_{i} = 0 \\ \\
p_{01}\mathcal{N}\left ( 0,\sigma ^{2}_{0} \right )+(1-p_{01})\mathcal{N}\left ( 1,\sigma ^{2}_{0} \right ) & \textrm{if} \quad  s_{i} = 1
\end{cases}
\end{equation}
In order to find the global maximum of (\ref{equ_13}) we decrease the variance $\sigma^{2}_{0}$ in each iteration of the algorithm gradually, which averts the local maximum of 
(\ref{equ_13}). Although we have converted the discrete variables $s_{i}$ to the continuous form, finding the optimal value of $\mathbf{s}$ using (\ref{equ_13}) is still complicated. Thus, we
propose an algorithm that estimates the unknown original signal $\mathbf{w}$ by estimating its components ($\mathbf{s}$ and $\boldsymbol{\theta}$ in 
(\ref{equ_vec})) iteratively. We follow a two-step approach to estimate the $\mathbf{w}$ vector. In the first step, we estimate the amplitude vector $\boldsymbol{\theta}$ (i.e., $\widehat{\boldsymbol{\theta }}$) based 
on the known estimation of $\mathbf{s}$ (i.e., $\widehat{\mathbf{s}}$) vector and the mixing observation vector $\mathbf{y}$. We call this expectation step (E-step). 


Having assumed the Gamma distribution as hyperpriors over the hyperparameters $\left \{ \gamma _{i} \right \}$ as explained in Section \ref{M_E_theta}, the following equation similar to (\ref{equ_23_1}) and
(\ref{equ_23_2}) can be derived as

\begin{align}
\label{equ_35_1}
\widehat{\boldsymbol{\theta }}&=\left ( \mathbf{\widehat{\Psi} }^{T}\mathbf{\widehat{\Psi} } +\sigma^{2}_{n}\mathbf{\Sigma }_{0} \right ) ^{-1}\mathbf{\widehat{\Psi} 
}^{T}\mathbf{y}\\
\label{equ_35_2}
&=\mathbf{\Sigma }^{-1}_{0} \mathbf{\widehat{\Psi}}^{T}\left ( \mathbf{\widehat{\Psi}}\boldsymbol{\Sigma }^{-1}_{0}\mathbf{\widehat{\Psi}}^{T}+\sigma^{2} _{n}\boldsymbol{I}_{N} 
\right )^{-1}\mathbf{y}
\end{align}
where $\boldsymbol{\widehat{\Psi}}=\mathbf{\Phi}\mathbf{\widehat{S}}$.

We call the second step of our approach maximization step (M-step). In this step, we find the estimate of $\mathbf{s}$ with the assumption of known vector 
$\widehat{\boldsymbol{\theta }}$ and the observation vector $\mathbf{y}$. Therefore, we can write the MAP estimate of $\mathbf{s}$ as

\begin{displaymath}
\mathbf{\widehat{s}}_{\mathrm{MAP}}=\argmax_\mathbf{s}
p(\mathbf{s}\mid\mathbf{y},\boldsymbol{\widehat{\theta}})\equiv
\argmax_\mathbf{s}p(\mathbf{s}\mid\boldsymbol{\widehat{\theta}})p(\mathbf{y}\mid\mathbf{s},\boldsymbol{\widehat{\theta}})\equiv
\end{displaymath}
\begin{equation}
\label{equ_36}
\argmax_\mathbf{s}p(\mathbf{s})p(\mathbf{y}\mid\mathbf{s},\boldsymbol{\widehat{\theta}})
\equiv\argmax_\mathbf{s}(\log (p(\mathbf{s}))+\log
(p(\mathbf{y}\mid\mathbf{s},\boldsymbol{\widehat{\theta}}))).
\end{equation}
Using (\ref{equ_31}) and (\ref{equ_32}), we can express $p(\mathbf{s})$ as a continuous variable

\begin{equation}
\label{equ_37}
\begin{split}
p\left ( \mathbf{s} \right )&=p\left ( s_{1} \right )\prod_{i=1}^{M-1}p\left ( s_{i+1}\mid s_{i} \right )\\
&=
p\exp\left ( \frac{-s^{2}_{1}}{2\sigma _{0}^{2}} \right )+\left ( 1-p \right )\exp\left ( \frac{-\left ( s_{1}-1 \right )^{2}}{2\sigma _{0}^{2}} \right )\\
&\quad\times \prod_{i=1}^{M-1}\left \{\vphantom{\exp\left ( \frac{-\left ( s_{i+1}-1 \right )^{2}}{2\sigma _{0}^{2}} \right )}\left [ p_{01}+\left ( 1-p_{10} \right ) \right ]\exp\left ( \frac{-s_{i+1}^{2}}{2\sigma _{0}^{2}} \right )\right.\\
&\quad+
\left.
\left [ p_{10}+\left ( 1-p_{01} \right ) \right ]\exp\left ( \frac{-\left ( s_{i+1}-1 \right )^{2}}{2\sigma _{0}^{2}} \right )\right \}.
\end{split}
\end{equation}
It remains to calculate $p\left ( \mathbf{y}\mid\mathbf{s},\widehat{\boldsymbol{\theta }} \right )$ in (\ref{equ_36}). From (\ref{equ_18}), we can write

\begin{equation}
\label{equ_38}
p\left ( \mathbf{y}\mid\mathbf{s},\widehat{\boldsymbol{\theta }} \right )=\frac{1}{\left (  \sqrt{2\pi\sigma ^{2}_{n}}\right )^{M}}\exp\left ( -\frac{\left \| \mathbf{y}-\boldsymbol{\Psi}\widehat{\boldsymbol{\theta}} \right \|^{2}_{2}}{2\sigma ^{2}_{n}} \right )
\end{equation}
where $\boldsymbol{\Psi}=\mathbf{\Phi}\mathbf{S}$. After calculating the two summands in (\ref{equ_36}), we can express the M-step as

\begin{equation}
\label{equ_39}
\widehat{\boldsymbol{\mathbf{s}}}=\argmax_\mathbf{s}\mathcal{L}\left ( \mathbf{s} \right )
\end{equation}
where

\begin{equation}
\label{equ_40}
\mathcal{ L}(\mathbf{s})=
\log \left (p\left ( s_{1} \right )\right)+\sum_{i=1}^{M-1}\log(p\left ( s_{i+1}\mid s_{i} \right ))-\frac{\left \| \mathbf{y}-\boldsymbol{\Psi}\widehat{\boldsymbol{\theta}} \right \|^{2}_{2}}{2\sigma_n^2}
\end{equation}
We can find the optimal solution of (\ref{equ_39}) by performing the steepest-ascent method. The expression for obtaining the sequence of optimal solutions in this method can be 
given as

\begin{equation}
\label{equ_41}
\mathbf{s}^{\left ( k+1 \right )}=\mathbf{s}^{\left ( k\right )}+\mu\frac{\partial
\mathcal{L}(\mathbf{s})}{\partial \mathbf{s}}
\end{equation}
where $\mu$ is the step size of the steepest-ascent method. The gradient term in (\ref{equ_41}) can be expressed in a closed form (see Appendix \ref{app_A}). Therefore (\ref{equ_41}) can be
rewritten as

\begin{equation}
\label{equ_42}
\mathbf{s}^{\left ( k+1 \right )}=\mathbf{s}^{\left ( k\right )}+\frac{\mu}{\sigma_0^2}
\mathbf{g}(\mathbf{s})
+\frac{\mu}{\sigma_n^2}\diag(\mathbf{\Phi}^T(\mathbf{\Psi}\boldsymbol{\widehat{\theta}}-\mathbf{y}))\cdot \boldsymbol{\widehat{\theta}}
\end{equation}
where $\mathbf{g}(\mathbf{s})$ which depends on $\sigma_{0}$ is derived in Appendix \ref{app_A}. 
Note that in the computation we decrease $\sigma_{0}$ in the consecutive iterations to guarantee the global maxima of (\ref{equ_40}). Hence, for each iteration we have
$\sigma_0^{(k+1)}=\alpha\sigma_0^{(k)}$, where $\alpha$ is selected in the range $\left [ 0.6,1 \right ]$. As the step size $\mu$ has a great effect on the convergence of the
Block-IBA, its proper range is analytically determined in Section \ref{sec_converge} to guarantee the convergence of Block-IBA (with a probability close to one). If the columns of $\mathbf{\Phi}$ are normalized to have unit norms, the range for step size $\mu$ can be expressed as

\begin{equation}
\label{equ_43}
0<\mu<\frac{2}{\frac{1}{\sigma_0^2}+\frac{M{M^{*}}^2}{\sigma_n^2}}
\end{equation}
where $M^{*}=\sigma_\theta Q^{-1}(\frac{1-\sqrt[M]{0.99}}{2})$ and $Q^{-1}(\cdot)$ is the inverse Gaussian Q-function. 

We initialize the proposed Block-IBA with
the minimum \ltwo-norm solution and use a decreasing threshold (i.e., $Th^{(k+1)}=\alpha Th^{(k)}$) so that the sampling matrix $\mathbf{\Phi}$ maximally samples the nonzero
elements of signal $\mathbf{w}$. In fact, the value of $Th$ optimally selects the number of nonzero elements in $\mathbf{s}$ vector.

Finally, it is observed from (\ref{equ_42}) that the second
summand controls the block sparsity of the $\mathbf{w}$ vector, whereas the third summand controls the noise power, i.e. $||\mathbf{y}-\mathbf{\Phi}\mathbf{w}||^2_2$. For 
instance, when the value of $\sigma_n$ is much smaller than the value of $\sigma_0$, the third summand dominates the second summand and the block sparsity of
$\mathbf{w}$ is doomed while the optimal solution satisfies $\mathbf{y}=\mathbf{\Phi}\mathbf{w}$. However, to obtain a meaningful solution in terms of block sparsity and noise power
we should select appropriate values for $\sigma_0$ and  $\sigma_n$ which are comparable to each other.

In this section, we have presented the first and the second steps of the Block-IBA, i.e. E-step and M-step given in (\ref{equ_35_2}) and (\ref{equ_42}), respectively. In the
next section, we complete the Block-IBA by estimating the unknown parameters of the signal model in Section \ref{sec_sigmodel}.

\section{Learning The Signal Model Parameters}\label{sec_params}
 
The signal model presented in Section \ref{sec_sigmodel} is characterized by Markov chain parameters $p$ and $p_{01}$, the variance parameter of amplitudes $\sigma^2_\theta$, and 
the the AWGN variance $\sigma^2_n$. It is likely that some or all of these parameters will require tuning to obtain better estimate of the unknown original signal. For this purpose, we
develop some estimation algorithms which work together with Block-IBA in Section \ref{sec_Block-IBA} to learn all of the model parameters iteratively from the available data.

To obtain an estimate of $\sigma_\theta$, the method of moments estimator is appealing. This is because this estimator is easy to calculate and simple to implement. Moreover, this
estimator is useful if the data vector record is sufficiently long. It is observed from (\ref{equ_7}) that the samples of the unknown original signal $\mathbf{w}$ have the special form of 
Gaussian mixture (Bernoulli-Gaussian) PDF. Hence, it can be shown that the second moment of the samples of $\mathbf{w}$ vector can be given as

\begin{equation}
\label{equ_44}
\mathbb{E}(w_i^2)=(1-p)\sigma_\theta^2
\end{equation}
 We assume that the $\mathbf{\Phi}$ matrix has the columns with the unit norms and its 
elements have a uniform distribution between [-1,1]. From (\ref{equ_1}), we know that $y_{j}=\sum_{i=1}^{M}\phi _{ji}w_{i}+n_{j}$, and by neglecting the noise power we have

\begin{equation}
\label{equ_45}
\mathbb{E}\left ( y_{j}^{2} \right )=M\mathbb{E}\left ( \phi_{ji}^{2}  \right )\mathbb{E}\left ( w_{i}^{2} \right )
\end{equation}
Moreover, we know that $\sum_{j=1}^{N}\phi _{ji}^{2}=1$, hence
$\mathbb{E}\left ( \phi_{ji}^{2}  \right )=1/N$. Finally, from (\ref{equ_44}) and (\ref{equ_45}) we can obtain a simple update for $\sigma_\theta$ as

\begin{equation}
\label{equ_46}
\sigma _{\theta}^{\left ( k+1 \right )}=\sqrt{\frac{N\mathbb{E}(y_j^2)}{M(1-p^{\left ( k+1 \right )})}}
\end{equation}

In Section \ref{M_E_theta}, we have derived an update in (\ref{equ_30}) for $\beta\triangleq\sigma^{-2}_{n}$ when we assumed the Gamma distribution as hyperprior over 
the hyperparameter $\boldsymbol{\gamma}$ and 
the Gamma prior for the inverse of the noise
variance $\sigma^{2}_{n}$. 
Moreover, using the MAP estimation
method we can express the following update equations for the rest of parameters

\begin{equation}
\label{equ_48}
p^{\left ( k+1 \right )}= \frac{\left \| \mathbf{s} \right \|_{0}}{M}
\end{equation}

\begin{equation}
\label{equ_49}
p^{\left ( k+1 \right )}_{01}= \frac{\sum_{i=1}^{M-1}s_{i}\left ( 1-s_{i+1} \right )}{\sum_{i=1}^{M-1}s_{i}}
\end{equation}
A complete derivation of the update in (\ref{equ_48}) can be found in \cite{refe_21}, while the derivation of (\ref{equ_49}) is presented in Appendix \ref{app_B}. Before starting the estimate of unknown parameters
in each iteration, it is essential to first initialize the parameters at reasonable values. For instance, the initial value for $p^{\left ( 0 \right )}$ is an arbitrary value between 0.5 and 1.
Furthermore, the initial value for $\sigma_{n}$ can be given as $\sigma ^{\left ( 0 \right )}_{n}= \sqrt{\frac{\sum_{i=1}^{N}y_{i}^{2}-\left [ \mathbb{E}\left ( \mathbf{y} \right ) \right ]^{2}}{N}}$.
When the initial value $p^{\left ( 0 \right )}$ is known, the amplitude variance $\sigma_{\theta}$ can also be initialized as $\sigma _{\theta}^{\left (0 \right )}=\sqrt{N\mathbb{E}(y_j^2)/{(M(1-p^{\left ( 0 \right )}))}}$. 

 Fig.~\ref{fig: alg}  provides a pseudo-code implementation of our proposed Block-IBA that gives all steps in the algorithm including E-step, M-step, and Learning 
Parameter-step. By numerical study, we empirically find that the threshold parameter should be $Th= 0.5$ to achieve reasonable performance. This is because the value of $Th$
specifies the number of nonzero elements in $\mathbf{s}$ vector. We will elaborate on this parameter in Section \ref{sec_experiment}.

\begin{figure}
  \centering \vrule
    \begin{minipage}{8.5cm} 
    \hrule \vspace{0.5em} 
    \begin{minipage}{7.5cm} 

      {
        \footnotesize
        \def\baselinestretch{1}

        \begin{itemize}
        \item Initialization:

           \begin{enumerate}
             \item Let $\boldsymbol{\Omega}_{0}$ equal to the initial parameter
             estimation:

             $p^{(0)}$:\quad arbitrary value in [0.5 1],\\
             $\sigma_{\theta}^{(0)}=\sqrt{\frac{NE(y_j^2)}{M(1-p^{(0)})}}$,
             $\sigma ^{\left ( 0 \right )}_{n}= \sqrt{\frac{\sum_{i=1}^{N}y_{i}^{2}-\left [ \mathbb{E}\left ( \mathbf{y} \right ) \right ]^{2}}{N}}$.

             \item Let $\mathbf{w}_0$, $\mathbf{s}_0$ and $\boldsymbol{\theta}_0$ equal to the initial solution from the minimum \ltwo-norm solution:

             $\mathbf{w}_0=\mathbf{\Phi}^T(\mathbf{\Phi}\mathbf{\Phi}^T)^{-1}\mathbf{y}$,

             $\mathbf{s}_0=(\mathbf{w}_0>Th)$, $\boldsymbol{\theta}_0=\mathbf{w}_0.(\mathbf{w}_0>Th)$.
           \end{enumerate}

        \item Until Convergence do:
          \begin{enumerate}
             \item E-step: solution obtained in (\ref{equ_35_1}) or (\ref{equ_35_2}).
             \item M-step:
             \begin{itemize}
             \item for $k=1,\dots,n_{\mathrm{iteration}}$:
                    \begin{itemize}
                       \item Update $\mathbf{s}$ with (\ref{equ_42})
                    \end{itemize}
                    \begin{itemize}
                       \item Update $\sigma_0^{(k+1)}=\alpha\sigma_0^{(k)}$
                    \end{itemize}
             \end{itemize}
             \item Parameter Estimation Step: using (\ref{equ_30}), (\ref{equ_46}), (\ref{equ_48}) and (\ref{equ_49}).

          \end{enumerate}
        \item Final answer is $\hat{\mathbf{w}}=\mathbf{s}_{\mathrm{final}}\boldsymbol{\theta}_{\mathrm{final}}$.
        \end{itemize}
      }
    \end{minipage}
    \vspace{1em} \hrule
  \end{minipage}\vrule \\
\caption{The overall Block-IBA
estimation.} \label{fig: alg}
\end{figure}

\section{Analysis of Global Maximum and Local Maxima}\label{sec_converge}

To ensure the convergence of Block-IBA, it is essential to examine the global maximum of the cost function 
$\mathcal{L}\left ( \mathbf{w} \right )\triangleq\log p\left ( \mathbf{w}\mid\mathbf{y} \right )$. Furthermore, as the steepest ascent is used in the M-step of Block-IBA,
it is necessary to analyze whether there is a global maximum for the cost function (\ref{equ_40}) which guarantees the convergence of the steepest ascent method. This analysis also
reveals the proper interval for the step size $\mu$. Finally, we show that there exist a unique local maxima for the cost function $\mathcal{L}\left ( \mathbf{w} \right )$ and this local
maxima is equal to the global maximum. Consequently, the convergence of the overall Block-IBA is guaranteed.

\subsection{Analysis of Global Maxima}\label{Sec_GlobalMax1}

The cost function $\mathcal{L}\left ( \mathbf{w} \right )$ which is called the log posterior probability function can be expressed as

\begin{equation}
\label{equ_52}
\mathcal{L}(\mathbf{w})\propto \log p(\mathbf{w})+ \log p(\mathbf{y}\mid\mathbf{w})
\end{equation}
Further manipulation of (\ref{equ_52}) gives

\begin{equation}
\label{equ_53}
\mathcal{ L}(\mathbf{w})=\sum_{i=1}^{M}\log(p\left ( w_{i}\right ))-\frac{\left \| \mathbf{y}-\boldsymbol{\Phi}\mathbf{w} \right \|^{2}_{2}}{2\sigma_n^2}
\end{equation}

\textit{Lemma 1}: The  cost function (\ref{equ_53}) is concave with respect to $\mathbf{w}$.

\textit{Outline of the Proof of Lemma 1}: As the proof is similar to the proof presented in Section V of~\cite{refe_21}, we only give an outline.

First, it is obvious that the quadratic function $L\left ( \mathbf{w} \right )\triangleq \left \| \mathbf{y}-\mathbf{\Phi }\mathbf{w} \right \|^{2}_{2}$ is convex. 
Then, it remains to prove the concavity of $\log(p\left ( w_{i}\right ))$. Finally, as the sum of concave functions is concave, the proof is completed. The PDF of the sources is
a Bernoulli-Gaussian hidden Markov model (BGHMM) given in (\ref{equ_7}). We can also rewrite (\ref{equ_7}) as $p(w_i)=\frac{p}{\sigma_1\sqrt{2\pi}}\exp(\frac{-w_i^2}{2\sigma_1^2})+\frac{1-p}{\sigma_2\sqrt{2\pi}}\exp(\frac{-w_i^2}{2\sigma_2^2})$,
where $\sigma_{1}$ is very small. It can be shown that the second derivative of $\log(p\left ( w_{i}\right ))$ is negative (see \cite{refe_21} for complete derivation) which results in
the concavity of $\log(p\left ( w_{i}\right ))$.

From the concavity of  (\ref{equ_53}), it can be concluded that the cost function $\mathcal{L}\left ( \mathbf{w} \right )$ has a unique global maxima.

\subsection{Analysis of Local Maxima}

In this section, we show that there exist a unique local maxima for the cost function $\mathcal{ L}(\mathbf{s})$ in (\ref{equ_40}), which in turn asserts that the M-step (steepest-ascent) 
converges to this maximum point. To this end, we provide the following lemma.

\textit{Lemma 2}: The sequence $\mathcal{ L}(\mathbf{s}^{\left ( k \right )})$ based on the cost function $\mathcal{ L}(\mathbf{s})$ is a monotonically increasing sequence.

\textit{Outline of the Proof of Lemma 1}: As the proof is the generalization of the proof presented in Section V of \cite{refe_21}, we only give an outline.

First from (\ref{equ_40}), we define the following expression:

\begin{equation}
\label{equ_55}
\begin{split}
\mathcal{ L}(\mathbf{s}^{\left ( k+1 \right )})-\mathcal{ L}(\mathbf{s}^{\left ( k \right )})=\\
\log\frac{p\left ( s_{1}^{\left ( k+1 \right )} \right )}{p\left ( s_{1}^{\left ( k \right )} \right )}+\sum_{i=1}^{M-1}\log\frac{p\left ( s^{\left ( k+1 \right )}_{i+1}\mid s_{i}^{\left ( k+1 \right )} \right )}{p\left ( s^{\left ( k \right )}_{i+1}\mid s_{i}^{\left ( k \right )} \right )}-\frac{1}{2\sigma_n^2}\\
\times[\wbb_{k+1}^T\Hb\wbb_{k+1}^T-\wbb_{k}^T\Hb\wbb_{k}^T-\bb(\wbb_{k+1}-\wbb_{k})]
\end{split}
\end{equation}
where $\wbb_{k+1}\triangleq\Sb_{k+1}\boldsymbol{\theta}_{k}$, $\wbb_{k}\triangleq\Sb_{k}\boldsymbol{\theta}_{k}$, $\Hb\triangleq\mathbf{\Phi }^{T}\mathbf{\Phi }$ and
$\bb=2\mathbf{y}\mathbf{\Phi }^{T}$. 
We can combine the first and the second summands in (\ref{equ_55}) and rewrite the expression as
\begin{equation}
\label{equ_56}
\begin{split}
\mathcal{ L}(\mathbf{s}^{\left ( k+1 \right )})&-\mathcal{ L}(\mathbf{s}^{\left ( k \right )})=
\sum_{i=0}^{M-1}\log\frac{p\left ( s^{\left ( k+1 \right )}_{i+1}\mid s_{i}^{\left ( k+1 \right )} \right )}{p\left ( s^{\left ( k \right )}_{i+1}\mid s_{i}^{\left ( k \right )} \right )}-\frac{1}{2\sigma_n^2}\\
&\times[\wbb_{k+1}^T\Hb\wbb_{k+1}^T-\wbb_{k}^T\Hb\wbb_{k}^T-\bb(\wbb_{k+1}-\wbb_{k})]
\end{split}
\end{equation}
It remains to show that (\ref{equ_56}) is positive. Recall that the M-step iterations are $\sbb_{k+1}=\sbb_{k}+\mu\cb_{k}$ where
 $\cb_k=\frac{\partial{\mathcal{L}(\sbb)}}{\partial{\sbb}}\Big|_{\sbb=\sbb_k}$. Consequently, we have $\wbb_{k+1}-\wbb_{k}=\mu\Cb_{k}\boldsymbol{\theta}_{k}$ where 
$\Cb_k=\diag(\cb_k)$. We substitute these equations in (\ref{equ_56}) and rewrite it as

\begin{equation}
\label{equ_57}
\begin{split}
\mathcal{ L}(\mathbf{s}^{\left ( k+1 \right )})&-\mathcal{ L}(\mathbf{s}^{\left ( k \right )})=
\sum_{i=0}^{M-1}\log\frac{p\left ( s^{\left ( k+1 \right )}_{i+1}\mid s_{i}^{\left ( k+1 \right )} \right )}{p\left ( s^{\left ( k \right )}_{i+1}\mid s_{i}^{\left ( k \right )} \right )}-\frac{\mu}{2\sigma_n^2}\\
&\times[\mu\boldsymbol{\theta}_{k}^T\Cb_k\Hb\Cb_k\boldsymbol{\theta}_{k}+(2\wbb_k^T\Hb-\bb)\Cb_k\boldsymbol{\theta}_{k}]
\end{split}
\end{equation}
After some algebra which is very similar to that presented in Appendix III in \cite{refe_21}, we obtain the following lower bound for (\ref{equ_57}) as

\begin{equation}
\label{equ_58}
\frac{1}{\mu}[\mathcal{L}(\sbb^{\left ( k+1 \right )})-\mathcal{L}(\sbb^{\left ( k\right )})]\geqslant\cb_k^T[\Ib-\mu\Rb-\frac{\mu}{2\sigma_n^2}\Rb_1\Hb\Rb_1]\cb_k
\end{equation}
where $\Rb_1=\diag(\theta_i)$ and $\Rb$ is defined as

\begin{equation}\nonumber
\Rb\triangleq\diag(\sum_{j=1}^2\frac{r_j(s^{(k)}_{i+1}\mid s^{(k)}_{i})}{2\sigma_0^2}))
\end{equation}
where $r_j(s^{(k)}_{i+1}\mid s^{(k)}_{i})\triangleq\frac{\pi_jg_j(s^{(k)}_{i+1}\mid s^{(k)}_{i})}{\sum_{j=1}^{2}\pi_jg_j(s^{(k)}_{i+1}\mid s^{(k)}_{i})}$, $g_j(s_{i+1}\mid s_i)=\frac{1}{\sigma_0\sqrt{2\pi}}\exp(\frac{-(s_i-m_j)^2}{2\sigma_0^2})$, 
$\pi_1\triangleq \left [ p_{01}+\left ( 1-p_{10} \right ) \right ]$ and $\pi_2\triangleq \left [ p_{10}+\left ( 1-p_{01} \right ) \right ]$ for $i=1,2,...,M-1$. Also, for $i=0$
$\pi_1\triangleq p$ and $\pi_2\triangleq 1-p$.
It can be shown that the symmetric matrix
$\Db\triangleq\Ib-\mu\Rb-\frac{\mu}{2\sigma_n^2}\Rb_1\Hb\Rb_1$ is Positive Definite (PD) if the step size $\mu$ is in the interval
(see complete proof in \cite{refe_21}).

\begin{equation}
\label{equ_mu}
0<\mu<\frac{2}{\frac{1}{\sigma_0^2}+\frac{M{M^{*}}^2}{\sigma_n^2}}
\end{equation}
where $M^{*}=\sigma_\theta Q^{-1}(\frac{1-\sqrt[M]{0.99}}{2})$ and $Q^{-1}(\cdot)$ is the inverse Gaussian Q-function. Therefore, the sequence
$\mathcal{ L}(\mathbf{s}^{\left ( k \right )})$ is a monotonically increasing sequence. As $\mathcal{ L}(\mathbf{s})$ is upper bounded by $M\log(p)$, the sequence has a limit and converges to this
local maximum point. Consequently, the convergence of M-step of steepest ascent method is guaranteed.

\subsection{Analysis of Global Maximum of overall Block-IBA}

To prove the existence of the global maximum for the proposed Block-IBA, we should prove that the log posterior probability sequence 
$\mathcal{L}\left ( \mathbf{w}^{\left ( k \right ) }\right )$ in (\ref{equ_53}) is an increasing sequence. Notice that this condition should be true in both E-step and M-step
of the Block-IBA. As the log posterior $\mathcal{L}\left ( \mathbf{w} \right )$ is equivalent to $\mathcal{L}\left ( \mathbf{s} \right )$ throughout the M-step, it is clear that
$\mathcal{L}\left ( \mathbf{w} \right )$ is increasing with respect to the sequence $\mathbf{w}^{\left ( k \right ) }$. Moreover, in E-step, the estimation of $\boldsymbol{\theta}$
is performed either through (\ref{equ_35_1}) or (\ref{equ_35_2}), which is the MAP estimation of amplitude vector $\boldsymbol{\theta}$. This MAP estimation implies the maximization
of the log posterior $\mathcal{L}\left ( \mathbf{w} \right )$. This is because the logarithm function is concave and monotonically increasing. Hence, the increasing characteristic of 
the log posterior $\mathcal{L}\left ( \mathbf{w} \right )$ is guaranteed in both E-step and M-step in each iteration. As a result, the sequence 
$\mathcal{L}\left ( \mathbf{w}^{\left ( k \right ) }\right )$ always converges to a local maxima $\wbb^{*}$. We have proved in \ref{Sec_GlobalMax1} that the 
$\mathcal{L}\left ( \mathbf{w}\right )$ is a concave function. Hence, this unique local maxima attained by the MAP estimate of block sparse sources in
Block-IBA is the global maximum.  

\section{Empirical Evaluation}\label{sec_experiment}

This section presents the experimental results to demonstrate the performance of Block-IBA. All the experiments are conducted for 400 independent simulation runs.
In each simulation run the elements of the matrix $\boldsymbol{\Phi}$ are chosen from a uniform distribution in [-1,1] with columns normalized to unit \ltwo-norm. The 
Block-sparse sources $\mathbf{w_{gen}}$ are synthetically generated using BGHMM in (\ref{equ_7}) which is based on Markov chain process. Unless otherwise stated,
in all experiments $p=0.9$, $p_{01}=0.09$ and $\sigma_{\theta}=1$ which are the parameters of BGHMM. The measurement vector
$\mathbf{y}$ is constructed by $\mathbf{y}=\mathbf{\Phi}\mathbf{w_{gen}}+\mathbf{n}$ where $\mathbf{n}$ is zero-mean AWGN with a variance tuned to a specified value of SNR
which is defined as 

\begin{equation}
\label{equ_SNR}
\mathrm{SNR\left ( dB \right )}\triangleq20\log_{10}\left ( \left \| \mathbf{\Phi}\mathbf{w_{gen}} \right \|_{2}/\left \| \mathbf{n} \right \|_{2} \right )
\end{equation}
In
addition, we utilize the following Normalized Mean Square Error (NMSE) as a performance metric

\begin{equation}\nonumber
\mathrm{NMSE}\triangleq\frac{\left \| \widehat{\mathbf{w}}-\mathbf{w_{gen}} \right \|^{2}_{2}}{\left \|\mathbf{w_{gen}} \right \|^{2}_{2}}
\end{equation}
where $\widehat{\mathbf{w}}$ is the estimate of the true signal $\mathbf{w_{gen}}$.

In the empirical studies, we compare the proposed Block-IBA with the following algorithms.

\begin{itemize}

\item EBSBL-BO and BSBL-EM, which are the two algorithms from the BSBL framework proposed in~\cite{refe_11}. In all the simulations, for EBSBL-BO, we set $h=4$ the block size parameter,
$noiseFlag =1$ (suitable for strongly noisy signal, i.g. $\mathrm{SNR}<20 \mathrm{dB}$), as suggested by the authors. For BSBL-EM,
we set  $h=4$ and devide the signal into equal block size in which the start of each block is known. Moreove, we set $LearnLambda=1$ for noisy cases.

\item BM-MAP-OMP, an algorithm proposed in~\cite{refe_10} where the sparsity pattern is modeled by Boltzmann machine. Throughout our experiments, we use the default
values for $k_{0}$, the prior belief on the average cardinality of the supports and $L$, the number of nonzero diagonals in the upper triangle part of the interaction matrix, as suggested by
the authors.

\item CluSS-MCMC, a hierarchical Bayesian model that uses Markov Chain Monte Carlo (MCMC) sampling approach, proposed in~\cite{refe_9}.
In all the experiments, we use all the default values suggested by the authors.

\item PC-SBL, an algorithm from a coupled hierarchical Gaussian framework proposed in~\cite{refe_12}. In all our experiments, we set $\beta=1$ the relevance parameter between
neighboring coefficient. Also, we used the 100 maximum number of iterations for the algorithm, as suggested by the authors.

\end{itemize}

\subsection{Performance of Block-IBA versus Block size}\label{seca}

Evidently, the strategy of selecting the blocks has a significant effect on the estimation performance. In this subsection, we examine the influence of the block size on the estimation
performance of Block-IBA where the block partition is unknown. To this end, we set up a simulation to compare the Block-IBA with all the other algorithms described above.
The size of matrix $\boldsymbol{\Phi}$ is $192 \times 512$, $\mathrm{SNR}=15 \mathrm{dB}$, and  $\sigma_{\theta}=1$. Based on the analytical result that the value of $\alpha$ should be in 
[$0.6$,$1$], we choose $\alpha=0.98$  in this experiment. The initial value of $\sigma_{0}$ is equal to 1. For these settings, the suitable interval of $\mu$ in (\ref{equ_mu}) is $0<\mu<2.1434\times 10^{-6}$. Hence, we select $\mu=10^{-6}$ and $Th=0.5$. Extensive experimental studies
demonstrate that for these parameters the $\mathcal{L}(\mathbf{s})$ converges to its maximum value within 4 or 5 iterations. Thus, 5 iterations are used for M-step. Also, we stop the
overall Block-IBA when the convergence criterion $\frac{||\widehat{\wbb}^{(k)}-\widehat{\wbb}^{(k-1)}||_2}{||\widehat{\wbb}^{(k)}||_2}<0.001$ is satisfied, where
$\widehat{\wbb}$ is the estimate of the true signal $\mathbf{w_{gen}}$ and $k$ is the iteration number.

\begin{figure}[tb]
\begin{center}
\includegraphics[width=8cm]{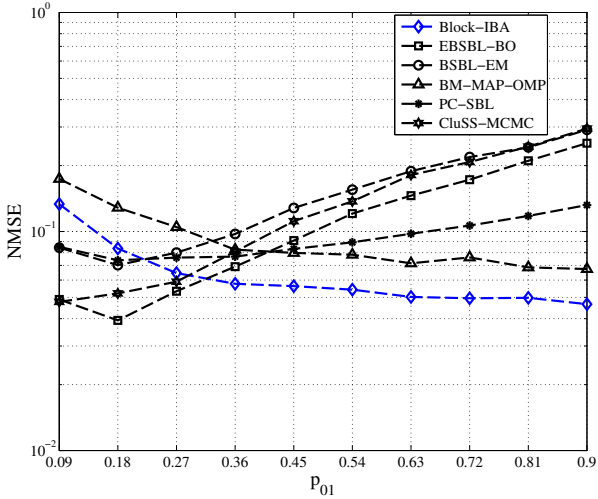}
\end{center}
\caption{\footnotesize NMSE versus $p_{01}$ for Block-IBA and other algorithms. The results are averaged over 400 trials.}
\label{fig2}
\end{figure}

Recall from Section \ref{sec_sigmodel} that the block size and the number of blocks of $\mathbf{w}$ are proportional to $1/p_{01}$. That is, when $p_{01}$ is small $\mathbf{w}$
comprises small number of blocks with big sizes and vice versa. Hence, we vary the value of $p_{01}$ between $0.09$ and $0.9$ to obtain the NMSE for various algorithms.
The results of NMSE versus $p_{01}$ is shown in Fig. \ref{fig2}. As seen from the figure, for $p_{01}\geq0.36$ Block-IBA
outperforms all other algorithms, whereas for $0.09\leq p_{01}<0.36$ most of the other algorithms outperform the Block-IBA. These two different performances are due to the different signal models used in Block-IBA and other algorithms.

\begin{figure}[tb]
\begin{center}
\includegraphics[width=8cm]{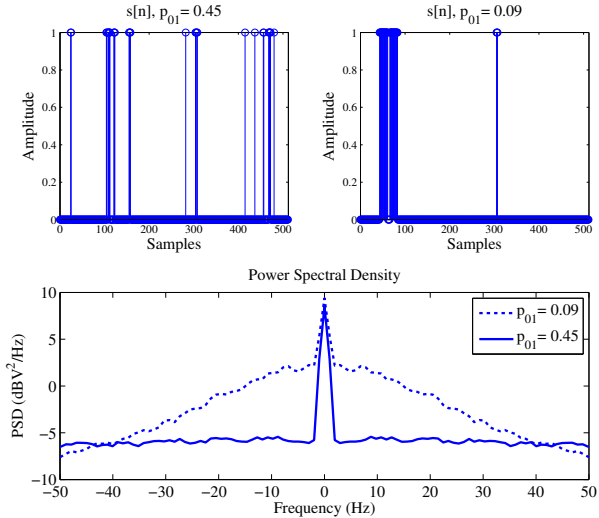}
\end{center}
\caption{\footnotesize Support vector $\mathbf{s}$ samples and the corresponding Power Spectral Density (PSD) for $p_{01}=0.09$ and $p_{01}=0.45$.}
\label{figpsd}
\end{figure}

Block-IBA uses BGHMM, hence it performs better when the block-sparse signal tends to be more non-i.i.d. Whereas all other algorithms use i.i.d. model, hence they perform better when  the block-sparse signal tends to be more i.i.d. For $0.09\leq p_{01}<0.36$, the support vector
$\mathbf{s}$ comprises a few number of blocks with large number of samples in each block as shown in Fig. \ref{figpsd} for $p_{01}=0.09$. As the elements of the $\mathbf{s}$ vector inside each block follow the first-order Markov chain process with two transition probabilities, they tend to be more i.i.d. when the number of samples is large.
This is obvious from Fig. \ref{figpsd} where the PSD of $\mathbf{s}$ vector shows wider 
bandwidth for $p_{01}=0.09$. On the other hand, for $p_{01}\geq0.36$  the support vector $\mathbf{s}$ consists of more blocks with fewer samples inside each block as shown in Fig. \ref{figpsd} for $p_{01}=0.45$.
Hence, the samples of $\mathbf{s}$ vector inside each block follow the first-order Markov chain process more accurately and they tend to be more non-i.i.d. This is seen from Fig. \ref{figpsd}
where PSD for $p_{01}=0.45$ is narrower compared to that for $p_{01}=0.09$.


\subsection{Performance of Block-IBA versus algorithm parameters}\label{sec_param}

The performance of the Block-IBA is affected by the parameters $\alpha$ and $Th$ (see Fig. \ref{fig: fig3-alpha} and Fig. \ref{fig: thfactor}). In this experiment, we use some simulations to examine the effects of these two parameters on the performance of Block-IBA.

\subsubsection{Performance versus parameter $\alpha$ }
\begin{figure}
\begin{center}
   \subfigure[]{
   \label{fig: alphanoise}
   \includegraphics[width=8cm]{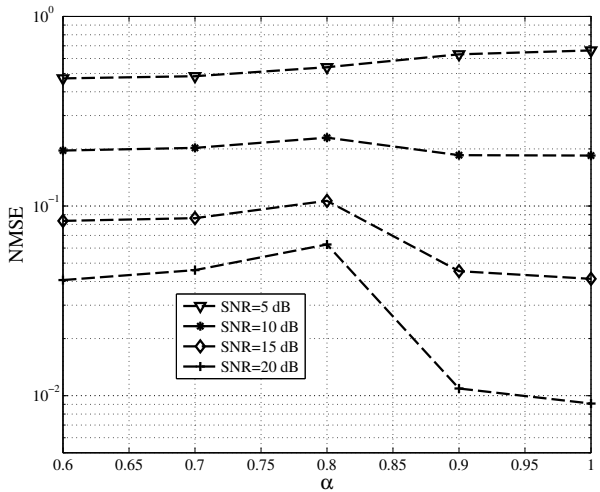}}
   \hskip 0.01\textwidth
   \subfigure[]{
   \label{fig: alphasparsity}
   \includegraphics[width=8cm]{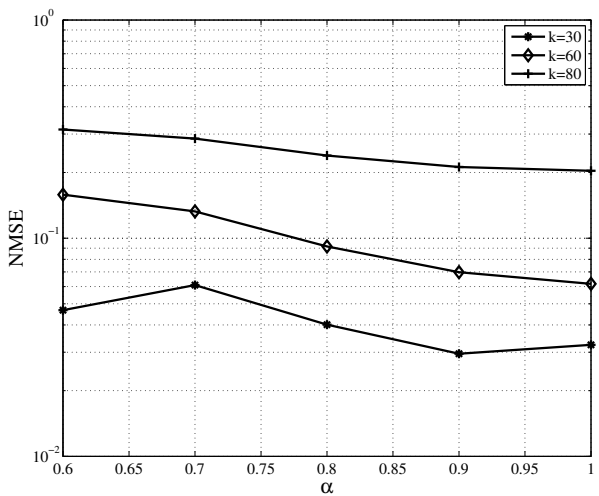}}
\end{center}
\caption{\footnotesize Performance of the Block-IBA vs parameter $\alpha$ for
 $N=192$, $M=512$ and $\sigma_\theta=1$. In
(a),  the number of active sources, $k$,  is fixed to 50 and the effect of $\mathrm{SNR}$ is investigated. In (b),
$\mathrm{SNR}$ is fixed to 15 dB and the effect of sparsity factor is
assessed. Values of $k$ are 30, 60, 80. Results are averaged over 400 simulations.}
 \label{fig: fig3-alpha}
\end{figure}

As discussed in Section \ref{sec_Block-IBA}, the parameter $\alpha$ controls the decay rate of $\sigma_{0}$ ($\sigma_0^{(k+1)}=\alpha\sigma_0^{(k)}$) to avert the local maxima of (\ref{equ_13}). To investigate its influence on the performance, we set up simulations to serve the NMSE versus parameter $\alpha$ for different values of $\mathrm{SNR (dB)}$ defined in (\ref{equ_SNR}) and 
the average number of active sources $k=\mathbb{E}\left [ \left | \mathcal{S} \right | \right ]=M(1-p)$. Note that $k$ also specifies the sparsity level of active sources. The results are illustrated in Fig. \ref{fig: fig3-alpha}, where Fig. \ref{fig: alphanoise} and Fig. \ref{fig: alphasparsity} represent the NMSE versus parameter $\alpha$ for different values of SNR and $k$, respectively.

As seen from Fig. \ref{fig: alphanoise}, the suitable range for the value of $\alpha$ is [0.9,1). In the experiment of Fig. \ref{fig: alphanoise}, we set $\mu=10^{-6}$, $k=50$, $M=512$ and the average number of nonzero blocks to 5 (i.e, $p_{01}=0.1$). Although the performance of Block-IBA
increases slightly when $\alpha$ is too close to one (e.g. $\alpha > 0.9$), it is observed that the performance shows little dependency on this parameter. Extensive simulation studies
show that $\alpha=0.98$ is an appropriate choice for block-sparse signal reconstruction. It can be shown that there is a unique sparsest solution for (\ref{equ_1}) when $k\leq N/2$
\cite{refe_25}-\cite{refe_26}. Therefore, in  Fig. \ref{fig: alphasparsity}, the results of NMSE versus parameter $\alpha$ for different values of sparsity levels ($30\leq k\leq 80$) are illustrated. It can be
seen that the Block-IBA still shows a low dependency on parameter $\alpha$ when sparsity level $k$ changes. In addition, the appropriate choice for $\alpha$ is in the range [0.9,1). Hence we chose $\alpha=0.98$.

\subsubsection{Performance versus threshold ($Th$) }

\begin{figure}[tb]
\begin{center}
\includegraphics[width=8cm]{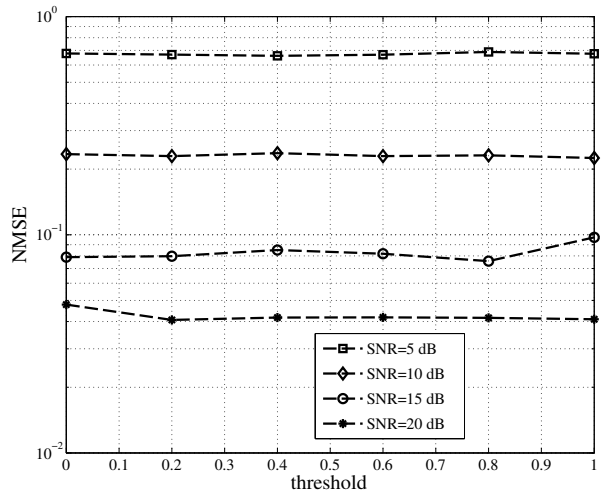}
\end{center}
\caption{\footnotesize The effect of the threshold on the Block-IBA. The
simulation parameters  are $M=512$, $N=192$, $p=.9$, $p_{01}=0.09$,$\sigma_{\theta}=1$ and $\mu=10^{-6}$. Results are averaged over 400
simulations.}
\label{fig: thfactor}
\end{figure}

In this experiment, we investigate the influence of the threshold parameter $Th$ on the performance of Block-IBA. We vary the value of $Th$ between $0$ and $1$
to obtain the NMSE versus $Th$ at different values of $\mathrm{SNR (dB)}$. The results are shown in Fig. \ref{fig: thfactor}. Although the performance of Block-IBA demonstrates a low dependency on
$Th$, extensive simulation studies shows that the optimal choice of $Th$ is $Th=0.5$.

\subsection{Effect of Sparsity Level on the Performance}

 Sparsity $\left |  \mathcal{S}\right |$ of the underlying signal is one of the key elements that has a considerable effect on any Compressed Sensing (CS) algorithm. For instance, the
capability of the algorithms to reconstruct the sparse sources can be determined by the level of the sparsity of the sources. That is, $N/2$ is the theoretical upper bound limit for 
maximum number of active sources of the signal to guarantee the uniqueness of the sparsest solution. However, most of the algorithms hardly achieve this limit in practice 
\cite{refe_25}. Hence, we can gain a lot of insight into an algorithm by manipulating this element and investigating the upcoming changes in the performance. To this end, in this
experiment, we study the performance of Block-IBA in terms of normalized sparsity ratio, $\eta \triangleq\frac{\mathbb{E}\left [ \left | \mathcal{S} \right | \right ]}{N}$.
For this experiment, the parameters of the signal model are set at $N=96$, $M=256$, $p_{01}=0.45$, $\sigma^{2}_{\theta}=1$, and $\mathrm{SNR}=15 \mathrm{dB}$. The value of
$p$ is set based on the specific value of $\eta$ and $p_{10}$ is set so that the expected number of active sources remains constant.

\begin{figure}[tb]
\begin{center}
\includegraphics[width=8cm]{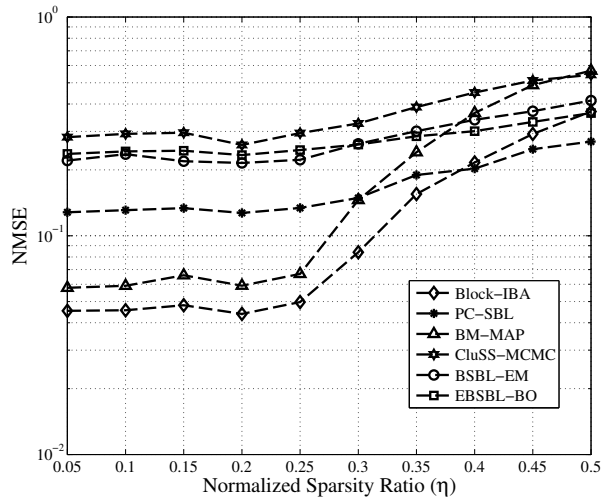}
\end{center}
\caption{\footnotesize NMSE vs. normalized sparsity ratio for different algorithms.
Simulation parameters are $M = 256$, $N = 96$, $\sigma_\theta = 1$,  $\mathrm{SNR}=15 \mathrm{dB}$ and $p_{01} = 0.45$. Results are averaged over 400 runs.}
\label{fig5}
\end{figure}

Fig. \ref{fig5} illustrates the resulting NMSE versus normalized sparsity ration ($\eta$) for various algorithms. The results are averaged over 400 trials. It is observed that, the
proposed Block-IBA presents the best performance among all the algorithms compared, for $\eta\leqslant 0.35$. For low sparsity level 
(e.g. $\eta\geq 0.4$) only PC-SBL outperforms Block-IBA.

\subsection{Real-World Data Experiment }

We have shown the effectiveness of the Block-IBA for recovering the synthetic data thus far. In this subsection, we evaluate the performance of Block-IBA for recovering an MRI image
\cite{refe_23}, \cite{refe_24}.
Images usually demonstrate the block-sparsity structures, particularly on over-complete basis such as wavelet or discrete cosine transform (DCT) basis. The coefficients of the image
in the wavelet or DCT domain tend to appear in clustered structures. Hence, images are appropriate data sets for testing the performance of block-sparse signal reconstruction 
algorithms. In this experiment, we consider an MRI image $\mathbf{I}$ of brain with the dimension of $256 \times 256$ pixels. To simulate MRI data acquisition process, the measurement matrix $\boldsymbol{\Phi}$ is obtained by the
linear operation $\mathbf{\Phi }=\mathbf{F_{1}}\mathbf{W}^{T}$. The first operation $\mathbf{W}^{T}$ is the 2-D and 2-level Daubechies-4 discrete wavelet transform (DWT) matrix. The second
operation, $\mathbf{F_{1}}$, is a 2-D partial discrete Fourier transform (DFT) matrix. In the experiment, we first randomly extract 216 rows from the 256 rows in the spatial frequency of the image
$\mathbf{I}$. Therefore, the partial DFT matrix $\mathbf{F_{1}}$ is a $216 \times 256$ compressed sensing matrix consisting of the
randomly selected 216 rows of the $256 \times 256$ DFT matrix. To reduce the computational complexity, we reconstruct the image $\mathbf{I}$ column by column. We compare the performance of Block-IBA with the other algorithms described in this section using the same 
parameter setups.

\begin{table}[htbp]
  \centering
  \caption{Performance of Block-IBA on MRI DataSet Compared to other algorithms}
    \begin{tabular}{|c|c|c|}
    \hline
    \textbf{Algorithm} & \textbf{NMSE (dB)} & \textbf{Runtime} \\
    \hline
    EBSBL-BO & -0.0518 & \textbf{27.09 sec} \\
    \hline
    BM-MAP-OMP & -2.37 & 22.36 min \\
    \hline
    CluSS-MCMC & -19.93 & 29.74 min \\
    \hline
    PC-SBL & -24.13 & 54.66 sec \\
    \hline
    BSBL-EM & -24.23 & 5.91 min \\
    \hline
    Block-IBA & \textbf{-26.78} & \textbf{75.39 sec} \\
    \hline
    \end{tabular}
  \label{tab:1}
\end{table}

 The performance of various algorithms is summarized in Table \ref{tab:1}. The Block-IBA outperforms all the other algorithms in respect of NMSE.
Although EBSBL-BO algorithm appears to be as the fastest algorithm, it shows a very poor performance. Considering Runtime with reasonable performance, only PC-SBL performs faster 
than Block-IBA. 
\begin{figure*}[tb]
\centering
{
\includegraphics[width=5cm]{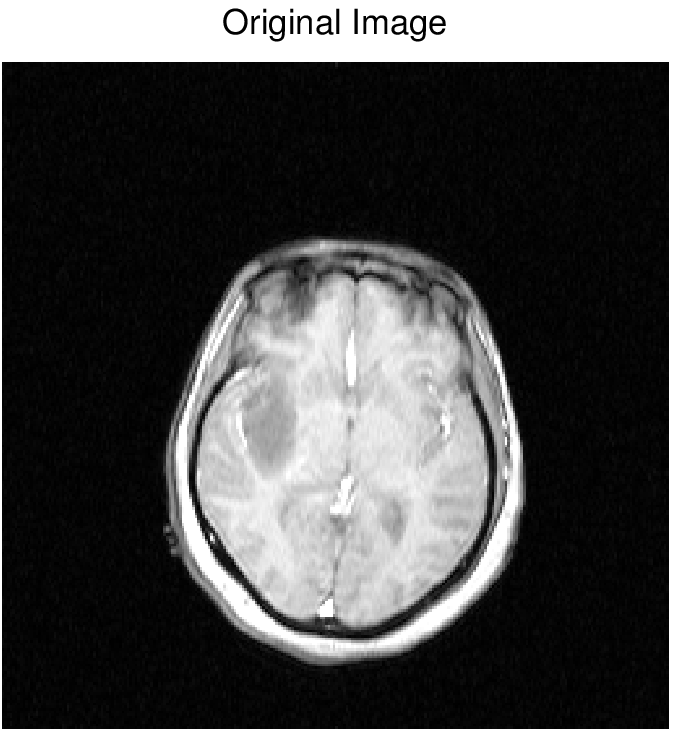}
\label{fig_ch:subfig1}
}
{
\includegraphics[width=5cm]{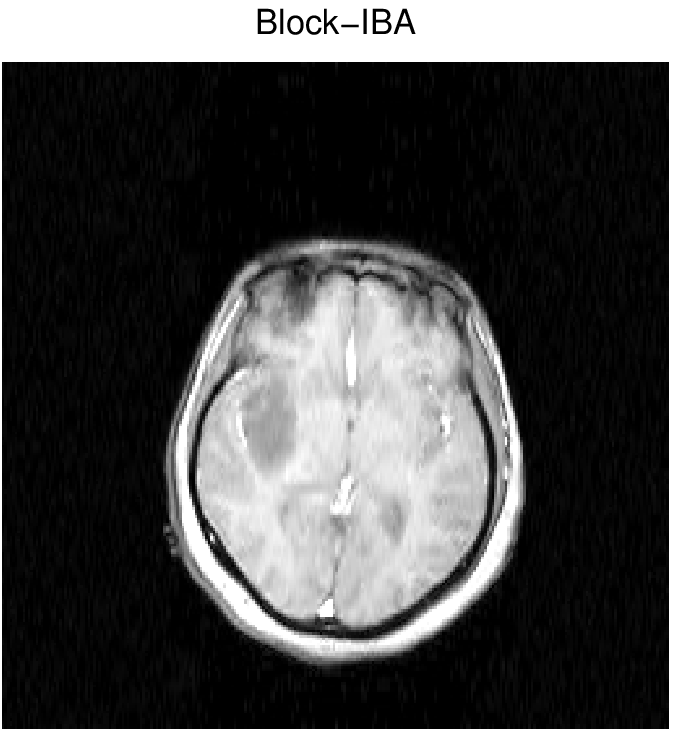}
\label{fig_ch:subfig2}
}
{
\includegraphics[width=5cm]{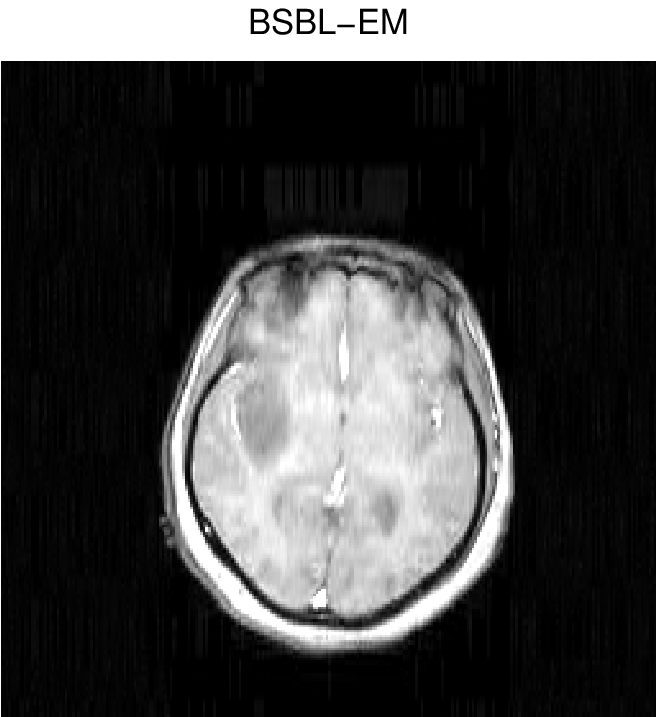}
\label{fig_ch:subfig2}
}
{
\includegraphics[width=5cm]{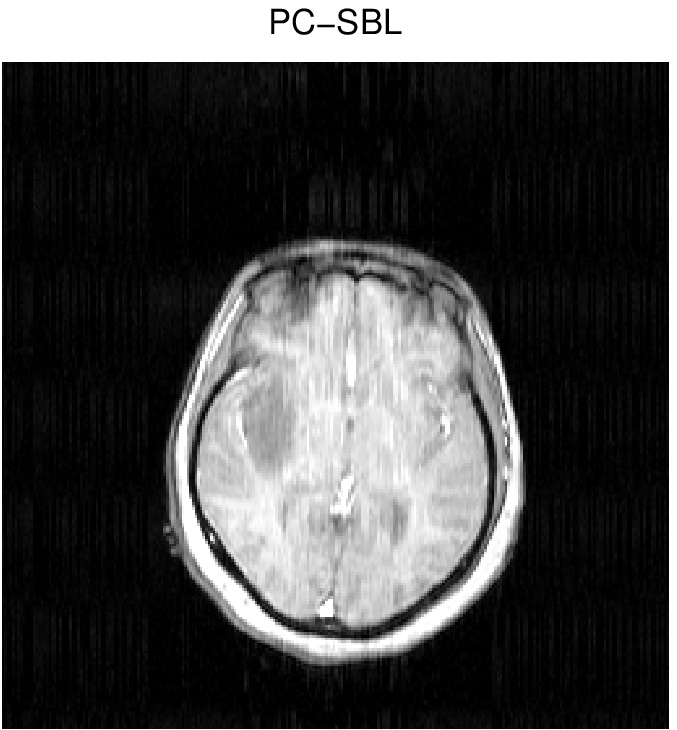}
\label{fig_ch:subfig3}
}
{
\includegraphics[width=5cm]{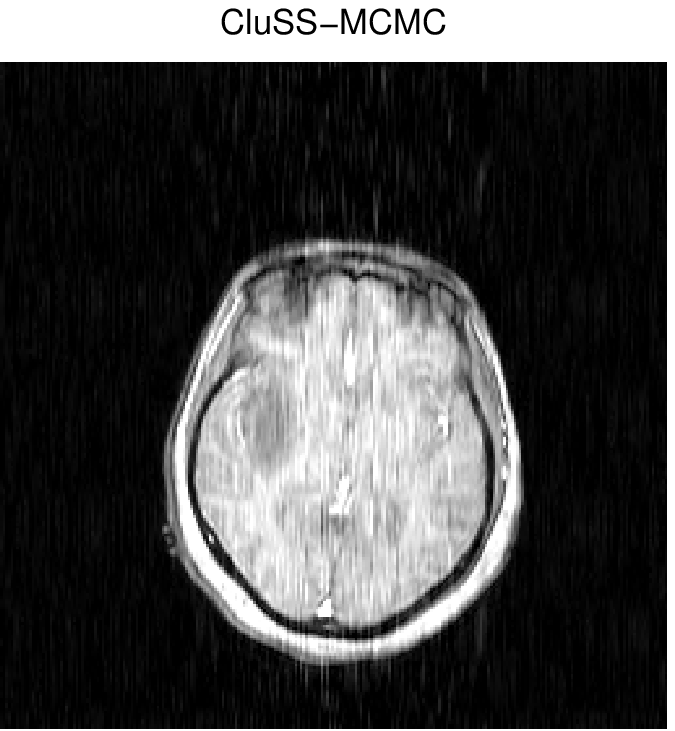}
\label{fig_ch:subfig4}
}
{
\includegraphics[width=5cm]{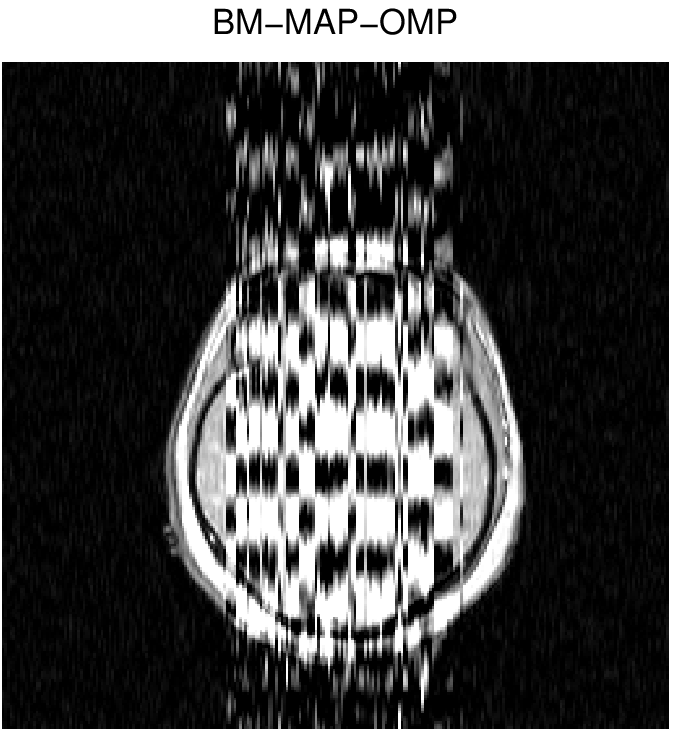}
\label{fig_ch:subfig4}
}
{
\includegraphics[width=5cm]{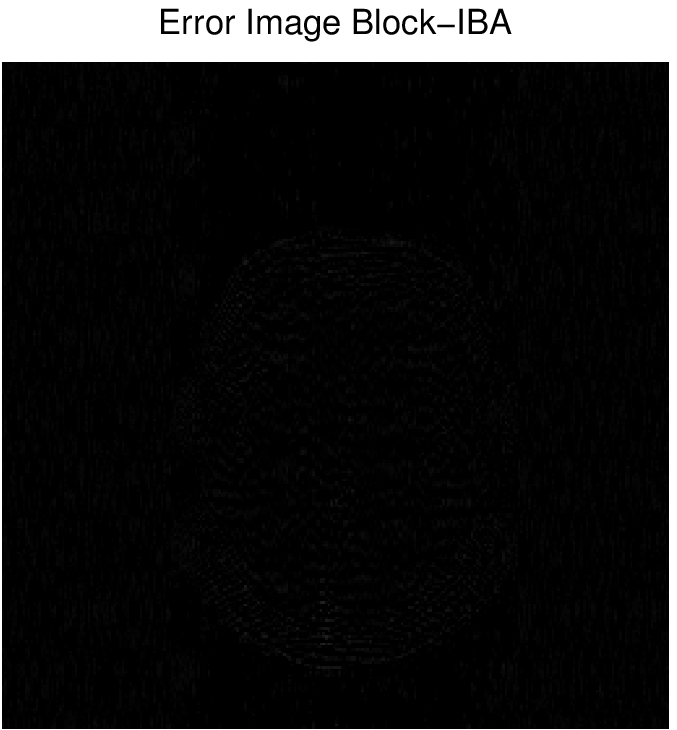}
\label{fig_ch:subfig4}
}
{
\includegraphics[width=5cm]{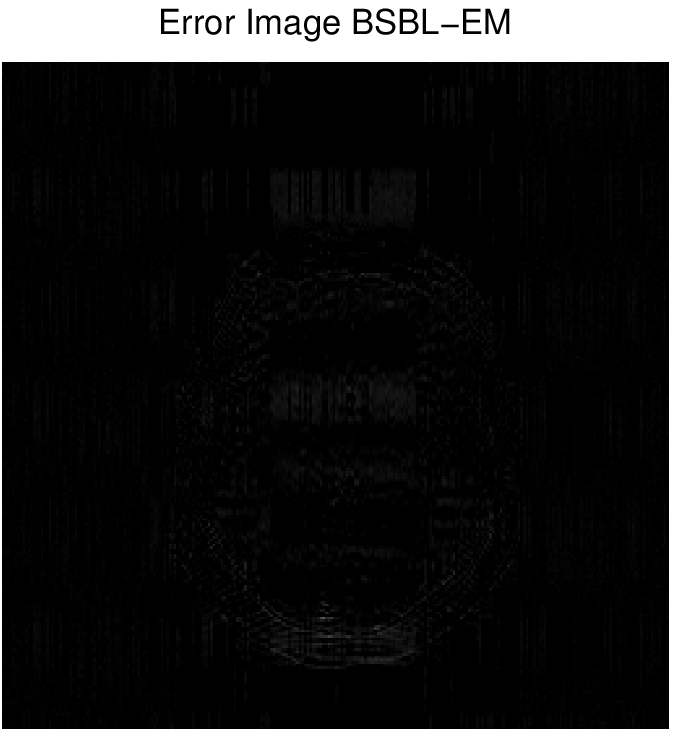}
\label{fig_ch:subfig4}
}
{
\includegraphics[width=5cm]{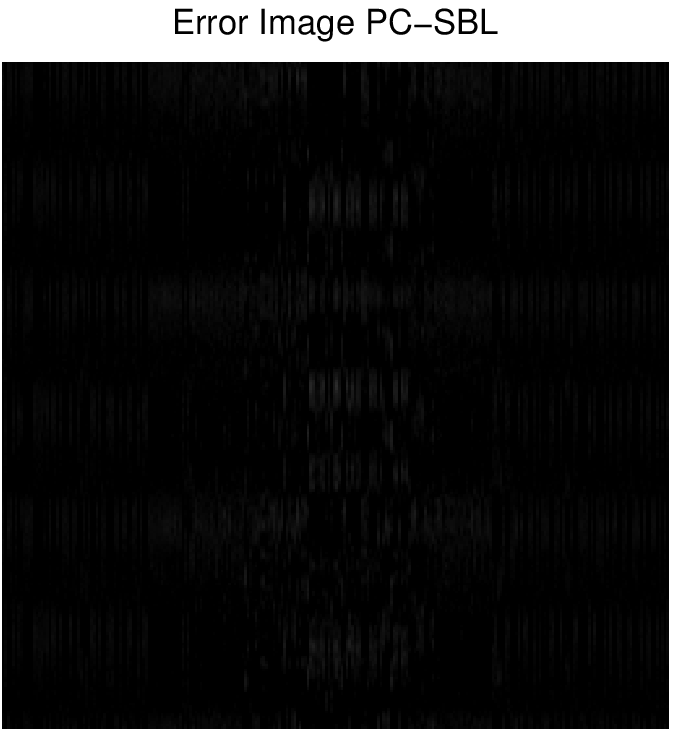}
\label{fig_ch:subfig4}
}
{
\includegraphics[width=5cm]{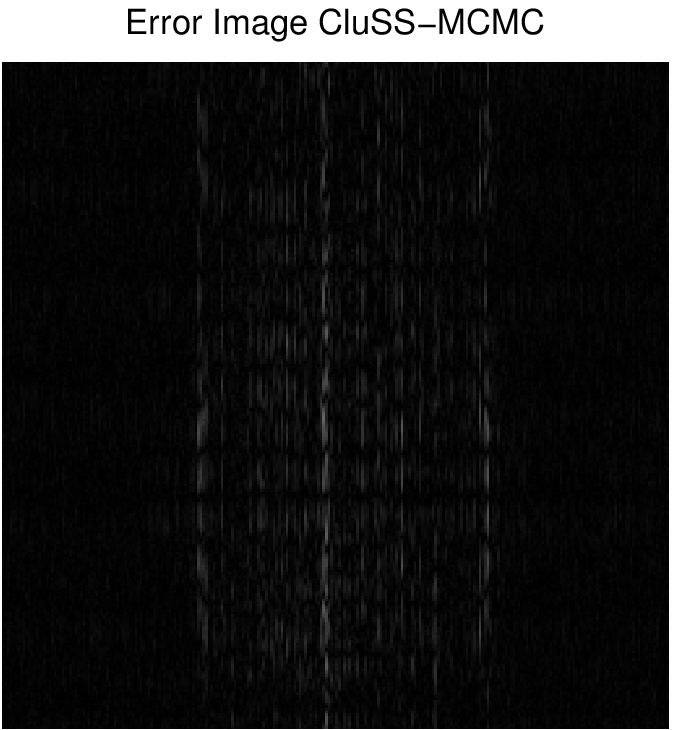}
\label{fig_ch:subfig4}
}
{
\includegraphics[width=5cm]{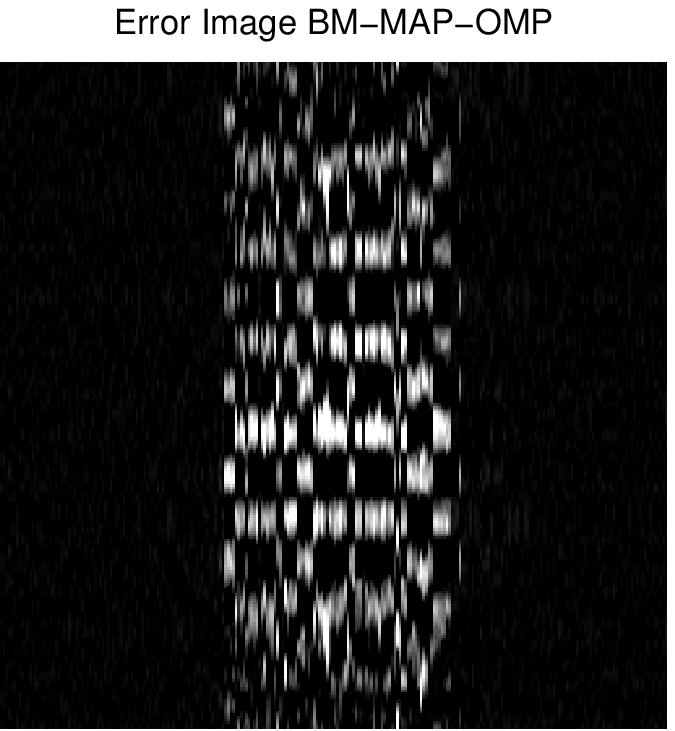}
\label{fig_ch:subfig4}
}
\caption{MRI image reconstruction performance for different algorithms accompanied by the corresponding error images.}
\label{fig3}
\end{figure*}
Figure \ref{fig3} compares the original MR image reconstructed by these algorithms and the corresponding error images. We have excluded the image reconstructed by EBSBL-BO algorithm because of its sever distortion. We observe that Block-IBA presents the best performance among all the algorithms.

\section{Conclusion}\label{sec_conc}
This paper has presented a novel Block-IBA to recover the block-sparse signals whose structure of block sparsity is completely unknown. Unlike the existing algorithms, we have modeled the cluster pattern of the signal using Bernoulli-Gaussian hidden Markov model (BGHMM), which better represents the non-i.i.d. block-sparse signals. The proposed Block-IBA utilizes adaptive thresholding to optimally select the nonzero elements of signal $\mathbf{w}$,  and takes the advantages of the iterative MAP estimation of sources and the EM algorithm to reduce the complexity of the Bayesian methods. The MAP estimation approach in Block-IBA renders learning all the signal model parameters automatically from the available data. We have optimized the M-step of the EM algorithm with the steepest-ascent method and provided an analytical solution for the step size of the steepest-ascent that guarantees the convergence of the overall Block-IBA. We have presented a theoretical analysis to show the global convergence and optimality of the proposed Block-IBA. Experimental results demonstrate that Block-IBA has a
low dependency on the algorithm parameters and hence is computationally robust. In empirical studies on synthetic data, the Block-IBA outperforms many state-of-the-art algorithms when the block-sparse signal comprises a large number of blocks with short lengths, i.e. non-i.i.d. Numerical experiment
on real-world data shows that Block-IBA achieves the best performance among all the algorithms compared at a very low computational cost.


%

\appendices
\section{Derivation of Steepest Ascent Formulation}\label{app_A}
The first derivative of (\ref{equ_40}) can be written as

\begin{equation}
\label{equ_59}
\begin{split}
\frac{\partial \mathcal{L}(\mathbf{s})}{\partial
\mathbf{s}}&=\frac{\partial}{\partial
\mathbf{s}} \log \left (p\left ( s_{1} \right )\right)+
\frac{\partial}{\partial
\mathbf{s}}\sum_{i=1}^{M-1}\log(p\left ( s_{i+1}\mid s_{i} \right ))\\
&\quad-\frac{1}{2\sigma_n^2}\frac{\partial}{\partial\mathbf{s}}(\mathbf{y}-\mathbf{\Phi }\mathbf{S}\boldsymbol{\widehat{\theta}})^T(\mathbf{y}-\mathbf{\Phi}\mathbf{S}\boldsymbol{\widehat{\theta }})
\end{split}
\end{equation}
Define $\mathbf{g}(\mathbf{s})\triangleq
-\sigma_0^2\frac{\partial}{\partial
\mathbf{s}}\log \left (p\left ( s_{1} \right )\right)-\sigma_0^2\frac{\partial}{\partial
\mathbf{s}}\sum_{i=1}^{M-1}\log(p\left ( s_{i+1}\mid s_{i} \right ))=\mathbf{{g}_1}(\mathbf{s})+\mathbf{{g}_2}(\mathbf{s})$ and 
$\mathbf{n\left ( \mathbf{s} \right )}\triangleq(\mathbf{y}-\mathbf{\Phi }\mathbf{S}\boldsymbol{\widehat{\theta}})^T(\mathbf{y}-\mathbf{\Phi}\mathbf{S}\boldsymbol{\widehat{\theta }})$. Then, the two scalar functions $g_{1}\left ( s_{1} \right )$ 
and $g_{2}\left ( s_{i+1} \right )$ ($i=1,2,\cdots,M-1 $) can be given as

\begin{equation}\nonumber
\label{equ_60}
g_1(s_1)=\frac{{ps_1}\exp(\frac{-s_1^2}{2\sigma_0^2})+{(1-p)(s_1-1)}\exp(\frac{-(s_1-1)^2}{2\sigma_0^2})
}{p\exp(\frac{-s_1^2}{2\sigma_0^2})+(1-p)\exp(\frac{-(s_1-1)^2}{2\sigma_0^2})}
\end{equation}

\begin{equation}\nonumber
\label{equ_61}
g_{2}(s_{i+1})=\frac{q_1s_{i+1}\exp(\frac{-s_{i+1}^2}{2\sigma_0^2})+{q_2(s_{i+1}-1)}\exp(\frac{-(s_{i+1}-1)^2}{2\sigma_0^2})
}{q_1\exp(\frac{-s_{i+1}^2}{2\sigma_0^2})+q_2\exp(\frac{-(s_{i+1}-1)^2}{2\sigma_0^2})}
\end{equation}
where $q_1=p_{01}+\left (1-p_{10}  \right )$ and $q_2=p_{10}+\left (1-p_{01}  \right )$. It can be shown that (see the complete proof in \cite{refe_21})

\begin{equation}
\label{equ_62}
\frac{\partial \mathbf{n}(\mathbf{s})}{\partial
\mathbf{s}}=2\cdot\diag(\mathbf{\Phi }^T\mathbf{\Phi }\mathbf{S}\boldsymbol{\widehat{\theta}}-\mathbf{\Phi }^T\mathbf{y})\cdot\boldsymbol{\widehat{\theta }}
\end{equation}
Therefore using (\ref{equ_62}), (\ref{equ_59}), (\ref{equ_41}) and the definitions of $\mathbf{g}(\mathbf{s})$ and $\mathbf{n\left ( \mathbf{s} \right )}$, the main steepest-ascent iteration in (\ref{equ_42}) can be obtained.
\section{MAP Update Equation for the Signal Model Parameter $p_{01}$}\label{app_B}
To calculate the update equation for parameter $p_{01}$, we use the MAP estimation approach, assuming the other parameters are known. Hence, we should maximize the
posterior probability $p\left (  p_{01}\mid \widehat{\mathbf{s}},\widehat{\boldsymbol{\theta }},\widehat{\sigma_{\theta } },\widehat{\sigma _{n}},\widehat{p},\mathbf{y}\right )$.
This probability is equivalent to $p\left ( \widehat{\mathbf{s}}\mid p_{01},p \right ) \cdot p\left ( \mathbf{y}\mid\widehat{\mathbf{s}} ,\widehat{\boldsymbol{\theta }},\widehat{\sigma _{n}}\right )$, where only $p\left ( \widehat{\mathbf{s}}\mid p_{01},p \right )$ depends on $p_{01}$. Therefore, the MAP estimate of parameter $p_{01}$ can be given as

\begin{equation}
\label{equ_63}
\begin{split}
 \widehat{p_{01}}_{MAP} &= \argmax_{p_{01}}p\left ( \widehat{\mathbf{s}}\mid p_{01},p \right )\\
&=\argmax_{p_{01}}p(s_{1})\prod^{M-1}_{i=1}p\left(s_{i+1}|s_{i}\right)
\end{split}
\end{equation}
where we have used the equation (\ref{equ_9}). As $p(s_{1})$ is independent of $p_{01}$, we can rewrite (\ref{equ_63}) as

\begin{equation}
\label{equ_64}
\begin{split}
 \widehat{p_{01}}_{MAP} &= \argmax_{p_{01}}\prod^{M-1}_{i=1}p\left(s_{i+1}|s_{i}\right)\\
&\equiv \argmax_{p_{01}}\sum_{i=1}^{M-1}\log\left ( p\left(s_{i+1}|s_{i}\right)\right )
\end{split}
\end{equation}
Define $\Gamma \triangleq \sum_{i=1}^{M-1}\log\left ( p\left(s_{i+1}|s_{i}\right)\right )$. Then, differentiating $\Gamma$ with respect to $p_{01}$ and using (\ref{equ_10}) gives

\begin{equation}
\label{equ_65}
\begin{split}
 \frac{\partial \Gamma }{\partial p_{01} }&=\sum_{i=1}^{M-1}\frac{\partial }{\partial p_{01}}\log\left ( p\left(s_{i+1}|s_{i}\right)\right )\\
&=\sum_{i=1}^{M-1}\left ( p_{01}^{-1}s_{i} \left ( 1-s_{i+1} \right )-s_{i}\left ( 1-p_{01} \right )^{-1}s_{i+1}\right )
\end{split}
\end{equation}

Equating (\ref{equ_65}) to zero and solving for $p_{01}$ result in the desired MAP update 

\begin{equation}
\label{equ_66}
p^{\left ( k+1 \right )}_{01}=\widehat{p_{01}}_{MAP}= \frac{\sum_{i=1}^{M-1}s_{i}\left ( 1-s_{i+1} \right )}{\sum_{i=1}^{M-1}s_{i}}
\end{equation}



\ifCLASSOPTIONcaptionsoff
  \newpage
\fi



%

\bibliographystyle{IEEEtran}
\bibliography{SPL}



%








\end{document}